\documentclass[a4paper,fleqn]{cas-dc}

\usepackage[authoryear,longnamesfirst]{natbib}

\usepackage{amsmath,amsfonts,amssymb}
\usepackage{amsthm}
\usepackage{subcaption}
\usepackage{algorithm}
\usepackage{algpseudocode}

\providecommand{\linkable}[1]{\url{#1}}

\begin{document}
\let\WriteBookmarks\relax
\makeatletter\setlength{\@fptop}{0pt}\makeatother

%% Float placement -------------------------------------------------
\renewcommand{\topfraction}{0.95}
\renewcommand{\bottomfraction}{0.95}
\renewcommand{\textfraction}{0.05}
\renewcommand{\floatpagefraction}{0.75}
\renewcommand{\dbltopfraction}{0.95}
\renewcommand{\dblfloatpagefraction}{0.75}
\setcounter{topnumber}{4}
\setcounter{bottomnumber}{4}
\setcounter{totalnumber}{6}
\setcounter{dbltopnumber}{4}
%% ---------------------------------------------------------------------

%% Running headers
\shorttitle{Repairing Shape-Prior Shortcuts in Long-Range Single-Shot FPP}
\shortauthors{A. Haroon et~al.}

\title[mode=title]{Repairing Shape-Prior Shortcuts in Long-Range Single-Shot Fringe Projection Profilometry}

%% Authors (Elsevier cas-dc frontmatter format). Lakshman dropped.
\author[1]{Adam Haroon}
\ead{aharoon@iastate.edu}
\credit{Conceptualization, Methodology, Software, Validation, Formal Analysis, Investigation, Data Curation, Writing -- Original Draft, Writing -- Review \& Editing, Visualization, Project Administration}

\author[1]{Cody Fleming}
\ead{flemingc@iastate.edu}
\credit{Supervision, Resources, Writing -- Review \& Editing}

\author[2]{Beiwen Li}
\cormark[1]
\ead{beiwen.li@uga.edu}
\credit{Supervision, Writing -- Review \& Editing}

\affiliation[1]{organization={Department of Mechanical Engineering, Iowa State University},
            addressline={2529 Union Drive},
            city={Ames},
            postcode={50011},
            state={Iowa},
            country={USA}}

\affiliation[2]{organization={College of Engineering, University of Georgia},
            addressline={597 D. W. Brooks Drive},
            city={Athens},
            postcode={30602},
            state={Georgia},
            country={USA}}

\cortext[1]{Corresponding author}

\begin{abstract}
Single-shot fringe projection profilometry (FPP) networks that regress depth
directly can exploit a shape-prior shortcut, recovering depth from object
boundaries rather than from fringe phase. On a photorealistic synthetic
benchmark (15{,}600 fringe images, 50 objects at 1.5--2.1~m standoff), the best
such UNet baseline plateaus at 14.54~mm object mean absolute error (MAE), and
neither more data nor more capacity removes the shortcut, because neither
changes the hypothesis space the optimizer searches. We introduce PhiCalNet,
which outputs a wrapped-phase representation $(\sin\phi, \cos\phi)$ and maps it
to depth through a fixed differentiable calibration layer, removing the
shape-prior solution architecturally rather than by a loss penalty. Because the
single-shot mapping is non-injective without fringe order, PhiCalNet takes the
fringe order as auxiliary input, an assumption a sensitivity analysis shows
tolerates realistic decoding error; a physics-informed (PINN) baseline with the
same physics as a soft penalty yields no gain, isolating the architectural
choice as the operative factor. PhiCalNet reduces object MAE 3.3$\times$ to
4.46~mm, its residual confined to 0.103\% of pixels at the $\pm\pi$ wrap
discontinuity, and a three-frame extension reaches 1.16~mm. Two checks agree:
interpretability makes phase the most decodable internal feature, and pixel-wise
conformal uncertainty quantification, to our knowledge the first for FPP,
localizes error at the same discontinuity, where rejecting the top 5\% of pixels
by snapshot disagreement cuts root-mean-square error by 64\% versus 3.5\% for
the baseline.
\end{abstract}

% \begin{highlights}
% \item PhiCalNet: a phase-intermediate architecture with phase-only output and a fixed differentiable calibration layer providing the deterministic phase-to-depth transform, yielding a 3.3$\times$ object-MAE reduction (4.46 vs.\ 14.54~mm) over a depth-regressing UNet baseline.
% \item A physics-informed neural network (PINN) counterfactual showing that loss-level physics enforcement on a depth-regressing network produces no measurable gain, isolating the architectural choice rather than the physics-in-the-loss as the operative factor.
% \item The residual error is localized to 0.103\% of pixels at the $\pm\pi$ wrap discontinuity, a structural limit of continuous regression of a discontinuous target; a three-frame extension reduces object MAE to 1.16~mm.
% \item The mechanistic interpretability profile of PhiCalNet inverts that of the UNet baseline across linear probing, dual-mode Grad-CAM, and a flat-plane out-of-distribution test, confirming phase-based rather than shape-prior decoding.
% \item First application of conformal prediction to FPP, with mechanistic interpretability and conformal uncertainty converging on the same wrap-discontinuity failure locus, offered as a diagnose-repair-verify template for physics-constrained learning.
% \end{highlights}


\maketitle

% -----------------------------------------------------------------------

\section{Introduction}
\label{sec:intro}

Single-shot fringe projection profilometry (FPP) promises dense, non-contact 3D measurement from one fringe image, and with it real-time operation under the non-stationary conditions (moving objects, perturbed sensors, variable lighting) that conventional FPP cannot tolerate. Conventional FPP reaches sub-millimeter accuracy in surface inspection~\citep{qian2021high,deng2016three}, robotic scanning~\citep{haroon2024autonomous,wang2024robotic}, and manufacturing process control~\citep{zhang2023machine,zhang2022systematic} through multi-step phase-shifting acquisition~\citep{zhang2016high,geng2011structured}, but that acquisition requires the scene, the camera-projector geometry, and the illumination to remain stable across a full pattern sequence, a constraint that learning-based single-shot reconstruction from one image~\citep{zuo2022deep,van2019deep,nguyen2020single,wang2021single,ikeda2025deep,li2025deep,zhu2022hformer,balasubramaniam2023single,wang2025end} removes. Single-shot methods have nonetheless been studied almost entirely at short range (below 1~m standoff); the long-range regime (standoff beyond 1~m) is both underexplored and intrinsically harder, because structured-light signal-to-noise ratio falls as $1/r^2$, fringe contrast drops as the projected pattern spreads, and the single-shot collapse of the phase-shifting pipeline is ill-posed without fringe-order information from a single image. We recap this ill-posedness in Section~\ref{sec:singleshot_theory} (established by \citet{haroon2026diagnosis}): the single-shot fringe-to-depth mapping is non-injective without fringe order, and the depth error induced by an incorrect fringe order grows as $Z^2$ in the working distance, so the regime becomes increasingly ill-posed with standoff.

The benchmark and the diagnosis that the present paper builds on are established by \citet{haroon2026diagnosis}. On FPP-ML-Bench~\citep{haroon2026fppml}, an open photorealistic synthetic benchmark built on the VIRTUS-FPP framework~\citep{HaroonVIRTUS2025} (15{,}600 fringe images, 300 depth reconstructions, 50 objects at 1.5--2.1~m standoff, with a standardized object/background/overall evaluation protocol), that work establishes a best depth-regressing UNet baseline at 14.54~mm object MAE, 18\% of the 80~mm object depth range, with only a 1.9$\times$ spread across four architectures. Three mechanistic interpretability probes (linear probing, Grad-CAM, and an in-range flat-plane out-of-distribution test) converge on the cause: the baseline solves the task via object-boundary shape priors rather than fringe-phase decoding. This is the shortcut-learning failure of Geirhos et al.~\citep{geirhos2020shortcut} instantiated in optical metrology, and the non-injectivity result recapped in Section~\ref{sec:singleshot_theory} identifies the mechanism, namely that the mapping the network is asked to learn does not uniquely exist without fringe-order information and an optimizer trained on $\ell_1$/$\ell_2$ regression resolves that non-injectivity by collapsing onto a shape-prior surrogate. Because the shortcut is a property of the hypothesis space the optimizer searches, additional data or larger models will not remove it. Section~\ref{sec:recap} recaps the pieces of this diagnosis that the present paper's comparisons require; the full treatment is in \citet{haroon2026diagnosis}.

We act on the diagnosis and verify the repair. We introduce PhiCalNet. Its operative design choice is not the calibration math itself (every phase-intermediate FPP network has classical triangulation calibration sitting somewhere in the pipeline) but a specific architectural framing: the network's output is wrapped phase $(\sin\phi, \cos\phi)$ rather than depth, and the only path from that output to the reconstructed depth runs through a fixed differentiable calibration layer that applies the FPP triangulation physics deterministically. The shape-prior solution available to a depth-regressing network is therefore removed from the hypothesis space architecturally rather than discouraged via a loss penalty, because the network never produces depth as a direct learned quantity. Consistent with the non-injectivity result recapped above, the fringe order that makes single-shot reconstruction well-posed is supplied to PhiCalNet as declared auxiliary side information, obtained from standard gray-code or multi-frequency decoding, rather than inferred from the single frame; the single fringe frame remains the only image input to the network. A physics-informed neural network (PINN) baseline that enforces the same physics as a soft loss penalty on an otherwise unconstrained depth-regressing network yields no measurable gain across a $\lambda$ sweep, isolating the architectural-versus-loss-level choice as the operative factor. PhiCalNet reduces single-shot object MAE by 3.3$\times$ over the UNet baseline (4.46 vs.\ 14.54~mm) on the same data, optimizer, and schedule. Its residual is carried by 0.103\% of object pixels at the $\pm\pi$ wrap discontinuity, a structural limit of continuous regression of a discontinuous target rather than a fitting failure; a three-frame extension reduces object MAE to 1.16~mm as the information-theoretic reading of the residual predicts, and a fringe-order sensitivity sweep shows the oracle fringe-order assumption used throughout is a convenient formalism rather than a strict deployment requirement.

We verify from two independent directions. The same MI battery applied to PhiCalNet inverts the UNet profile: phase becomes the most decodable internal feature (depth has no explicit internal representation, consistent with the calibration layer carrying the geometric transform), a dual-mode Grad-CAM analysis on the same trained weights shows fringe-favored attention at the layers driving the phase output with edge bias appearing only when gradients flow through the calibration layer, and the flat-plane test recovers a coherent $(\sin\phi,\cos\phi)$ field with the correct quadrature offset. Pixel-wise conformal uncertainty quantification using a snapshot ensemble, to our knowledge the first application of conformal prediction to FPP and one of the first to pixel-wise depth regression more broadly~\citep{angelopoulos2023conformal}, confirms the diagnosis from the outside: rejecting the top 5\% of object pixels by snapshot standard deviation reduces PhiCalNet RMSE by 64\% (20.6$\rightarrow$7.4~mm) while reducing the UNet baseline's RMSE by only 3.5\% (28.4$\rightarrow$27.4~mm). MI and UQ therefore converge on the same failure population, the $\pm\pi$ wrap discontinuity. We develop this convergence into a diagnose-repair-verify methodological template in Sections~\ref{sec:phicalnet_uq} and~\ref{sec:conclusion}, and treat it as a contribution beyond the FPP-specific result.

This paper makes the following contributions. First, we introduce PhiCalNet, a phase-intermediate architecture whose operative design choice is a phase-only output space coupled with a fixed differentiable calibration layer that maps phase to depth deterministically, and demonstrate via a PINN counterfactual that the architectural framing rather than the physics-in-the-loss is the operative factor. Second, we characterize the residual wrap-boundary error of PhiCalNet analytically and empirically, and show that a multi-frame extension closes the gap as the information-theoretic argument predicts and that a fringe-order sensitivity sweep bounds the accuracy required of an upstream fringe-order source. Third, we verify the repair via the same MI battery applied to PhiCalNet, which inverts the UNet profile across all three probes, with a dual-mode Grad-CAM analysis that isolates the calibration layer as the source of depth-output edge bias. Fourth, we apply conformal prediction to FPP for the first time and show that MI and UQ converge on a single geometric locus. The MI and UQ convergence at that locus is offered as a diagnose-repair-verify methodological template applicable to other physics-informed learning settings.

Section~\ref{sec:recap} recaps the FPP pipeline, the benchmark and baseline, the single-shot ambiguity theory, and the shape-prior diagnosis established by \citet{haroon2026diagnosis}. Section~\ref{sec:physics_aware} introduces PhiCalNet, its loss and a component ablation, results, and the PINN counterfactual. Section~\ref{sec:phaseunet_error_analysis} analyzes the residual wrap-boundary failure mode. Section~\ref{sec:phicalnet_multiframe} reports the multi-frame extension and fringe-order sensitivity. Section~\ref{sec:phicalnet_interpretability} verifies the repair mechanistically and Section~\ref{sec:phicalnet_uq} via uncertainty quantification. Section~\ref{sec:conclusion} concludes.

\section{Background: the Benchmark, the Ambiguity, and the Shape-Prior Diagnosis}
\label{sec:recap}

This section recaps only what the PhiCalNet repair and its verification depend on: the classical FPP pipeline whose stages PhiCalNet re-embeds architecturally (Section~\ref{sec:fpp_principles}), the benchmark and the depth-regressing baseline the repair is measured against (Section~\ref{sec:baseline}), the single-shot ambiguity theory that motivates routing reconstruction through phase (Section~\ref{sec:singleshot_theory}), and the mechanistic diagnosis of the UNet baseline that the PhiCalNet interpretability study inverts (Section~\ref{sec:interpretability}). The full benchmark, ablations, and diagnosis are developed in \citet{haroon2026diagnosis}; the treatment here is deliberately compact and is included so that the present paper stands on its own.

\subsection{FPP Principles}
\label{sec:fpp_principles}

FPP recovers per-pixel depth by projecting structured fringe patterns onto a scene and decoding the phase modulation that surface geometry imposes on those patterns~\citep{zhang2016high,geng2011structured}. The classical pipeline decomposes into three deterministic stages. First, a sinusoidal fringe pattern projected through a calibrated projector produces camera intensity images $I_n(x,y) = I'(x,y) + I''(x,y)\cos(\phi(x,y) + \delta_n)$, where $\phi(x,y) \in [-\pi,\pi)$ is the wrapped phase carrying depth information and $\delta_n = 2\pi n/N$ is the known phase shift of the $n$-th of $N$ patterns; acquiring the full $N$-step sequence recovers $\phi$ analytically. Second, the wrapped phase is ambiguous modulo $2\pi$, so the absolute phase
\begin{equation}\label{eq:prelim_absphase}
\Phi(x,y) = \phi(x,y) + 2\pi k(x,y)
\end{equation}
requires the integer fringe order $k(x,y) \in \mathbb{Z}$ at every pixel, recovered classically by an auxiliary procedure such as gray-code temporal unwrapping~\citep{sansoni1999three}. Third, with calibrated camera and projector models the absolute phase maps to a projector coordinate and the 3D point is recovered by triangulation~\citep{zhang2010recent}. Classical FPP achieves sub-millimeter accuracy because every stage is deterministic and well-posed. The single-shot setting collapses the $N$-step capture and the gray-code sequence into a single learned mapping from one fringe image to depth; Section~\ref{sec:singleshot_theory} formalizes the ambiguity this collapse introduces, and PhiCalNet (Section~\ref{sec:physics_aware}) re-embeds the wrapped-phase and calibration stages as fixed architectural components.

\subsection{Benchmark, Dataset, and Depth-Regressing Baseline}
\label{sec:baseline}

The experiments in this paper use FPP-ML-Bench~\citep{haroon2026fppml}, an open photorealistic synthetic benchmark for single-shot FPP generated with the VIRTUS-FPP framework~\citep{HaroonVIRTUS2025} in NVIDIA Isaac Sim. The dataset comprises 15{,}600 fringe images and 300 ground-truth depth maps across 50 object-diverse geometries scanned at 1.5--2.1~m standoff, partitioned 80/10/10 at the object level (240 training, 30 validation, 30 test object-viewpoints) so that evaluation is on unseen geometries. Each object-viewpoint yields 52 captured frames (18 horizontal and 18 vertical sinusoidal fringes, 14 gray-code frames, one black and one white); the single-shot input is the first horizontal fringe (frame 0), and the auxiliary fringe order $k_{\mathrm{gt}}$ used later by PhiCalNet is decoded from the accompanying 7-bit horizontal gray-code frames and median-filtered. Error is reported after denormalization on three pixel populations (object, background, and overall), because background pixels occupy 60--90\% of the image and otherwise dilute object error by nearly an order of magnitude.

\citet{haroon2026diagnosis} establishes, through systematic ablations on this benchmark, the design choices that govern single-shot performance: individual per-object depth normalization (which decouples object shape from absolute scale) improves object MAE 9.1$\times$ over raw depth; background fringes are signal rather than noise, and removing them degrades object MAE 2.8--7.3$\times$; homogeneous multi-frame training with phase-shifted frames of one orientation helps while mixing orientations hurts; and among six L1/L2 losses a Hybrid L1 with $\alpha=0.7$ (a masked object term plus a weak global regularizer) is best. Under this optimal configuration, a UNet~\citep{Ronneberger2015} with four encoder-decoder stages (31~M parameters, instance normalization) is the best of four architectures at 14.54~mm object MAE and 17.88~mm object RMSE, outperforming a transformer-bottleneck variant by 29\%, a residual UNet by 60\%, and a Pix2Pix conditional GAN by 91\%. The 1.9$\times$ spread across architectures and the 14.54~mm residual (18\% of the 80~mm object depth range) point to a representational rather than a capacity-bound limit. This UNet baseline, its optimal training configuration, and its Hybrid L1 loss are the reference against which PhiCalNet is measured throughout.

\subsection{Single-Shot Ambiguity and Long-Range Sensitivity}
\label{sec:singleshot_theory}

The single-shot formulation replaces the two analytical stages of Section~\ref{sec:fpp_principles} with a learned mapping $f_\theta: I_t \mapsto Z$. \citet{haroon2026diagnosis} prove that this mapping is ill-posed; we recap the two results the present paper depends on and refer the reader there for the formal statement, proof, and derivation. First, the single-shot fringe-to-depth mapping is non-injective without fringe-order information: the captured intensity depends on the absolute phase (Eq.~\eqref{eq:prelim_absphase}) through a $2\pi$-periodic cosine, so two absolute phases differing by an integer number of fringe periods produce identical intensities while corresponding to different projector coordinates and therefore different triangulated depths, and a single intensity image is consequently consistent with multiple physically distinct depths. Second, an incorrect fringe order produces a depth error that grows quadratically with the working distance. For a simplified triangulation model $Z = fb/d$ (focal length $f$, baseline $b$, disparity $d$), a fringe-order error of $m$ periods (fringe pitch $P$) induces
\begin{equation}\label{eq:range_error}
\left|\Delta Z\right| \propto \frac{Z^2}{fb}\,|mP|.
\end{equation}
The depth error caused by an incorrect fringe order therefore grows approximately quadratically with the object distance $Z$: fringe-order ambiguity exists at all working distances but its effect is significantly more severe in long-range FPP.

The implication for learning is direct. A network trained to regress $I_t \mapsto Z$ is asked to approximate a mapping that may not be uniquely defined, and under $\ell_1$ or $\ell_2$ regression it will learn an averaged or shortcut estimate rather than the physically correct depth. The principled response is to provide the information that resolves the ambiguity, so that the reconstruction pipeline becomes
\begin{equation}\label{eq:phase_pipeline}
I_t(x,y) \mapsto \bigl(\phi(x,y), k(x,y)\bigr) \mapsto \Phi(x,y) \mapsto Z(x,y),
\end{equation}
rather than a direct ill-posed map. PhiCalNet (Section~\ref{sec:physics_aware}) is one realization of this pipeline, and the residual-error analysis of Section~\ref{sec:phaseunet_error_analysis} localizes its residual to the $\pm\pi$ wrap-boundary pixels identified above as the locus of the underlying ambiguity, with the catastrophic-error magnitude there matching the one fringe period $\times\, Z^2/(fb)$ scaling of Eq.~\eqref{eq:range_error}.

\subsection{The Shape-Prior Diagnosis of the UNet Baseline}
\label{sec:interpretability}

\citet{haroon2026diagnosis} applies three mechanistic interpretability probes to the best UNet baseline and concludes that it solves the task via geometric shortcuts rather than fringe-to-depth physics. We recap the three results here because the PhiCalNet interpretability study (Section~\ref{sec:phicalnet_interpretability}) is defined as their inverse and refers to the baseline numbers directly.

\emph{Linear probing}~\citep{alain2016understanding} trains small auxiliary probes to predict a target from frozen intermediate activations; lower validation MSE means the target is more explicitly encoded. Across the nine UNet layers, geometric edge maps are decoded 2.82$\times$ more readily than depth values on average (Table~\ref{tab:probing_results}); although edges are more decodable than depth at every layer, the depth/edge gap narrows toward parity in the decoder (\texttt{dec2}--\texttt{dec3}), where depth is most decodable and where the network produces its output, indicating that a depth-as-shape representation is consolidated in the decoder.

\emph{Grad-CAM}~\citep{selvaraju2017grad} attribution, correlated against a Sobel edge map and a fringe-intensity-variation map, shows the network attends 1.28$\times$ more strongly to object boundaries than to fringe texture on average (Table~\ref{tab:gradcam_results}), with the preference strongest at the final decoder layer \texttt{dec4} (ratio 1.38).

\emph{Flat-plane out-of-distribution testing}~\citep{geirhos2020shortcut} presents a featureless flat plane at 1.8~m, within the trained depth range, that retains valid fringes but has no internal shape cues. The baseline predicts near-zero depth across most of the surface (mean 0.088 normalized units, range extending to $[-1.19, 4.38]$), effectively classifying the plane as background: encoder attention engages the fringes inside the rectangular region but the network cannot translate that fringe information into uniform depth without a shape template to fill in. (The corresponding input, prediction, and layer-wise Grad-CAM figures appear in \citealp{haroon2026diagnosis}.)

\begin{table}[pos=tbp]
\caption{UNet baseline linear probing validation MSE loss for geometry (edges) versus depth across the nine layers (recap from \citealp{haroon2026diagnosis}). Lower is better; edges are consistently more decodable than depth.}
\label{tab:probing_results}

\centering
\small
\resizebox{\ifdim\width>\columnwidth\columnwidth\else\width\fi}{!}{%
\begin{tabular}{@{}lccc@{}}
\toprule
\textbf{Layer} & \textbf{Edges Loss} & \textbf{Depth Loss} & \textbf{Ratio (Depth/Edges)} \\
\midrule
enc1 (skip1)    & 0.000387 & 0.002284 & 5.90 \\

enc2 (skip2)    & 0.000343 & 0.001765 & 5.15 \\

enc3 (skip3)    & 0.000336 & 0.001391 & 4.14 \\

enc4 (skip4)    & 0.000399 & 0.000888 & 2.23 \\

bottleneck      & 0.000468 & 0.001099 & 2.35 \\

dec1 (up1)      & 0.000439 & 0.000882 & 2.01 \\

dec2 (up2)      & 0.000375 & 0.000411 & 1.10 \\

dec3 (up3)      & 0.000308 & 0.000371 & 1.20 \\

dec4 (up4)      & 0.000355 & 0.000527 & 1.48 \\

\textbf{Average} & \textbf{0.000379} & \textbf{0.001069} & \textbf{2.82} \\
\bottomrule
\end{tabular}}
\end{table}

\begin{table}[pos=tbp]
\caption{UNet baseline Grad-CAM correlation analysis, attention to edges versus fringe patterns across layers (recap from \citealp{haroon2026diagnosis}). Ratio $>1$ indicates stronger attention to geometric boundaries than fringe patterns.}
\label{tab:gradcam_results}

\centering
\small
\resizebox{\ifdim\width>\columnwidth\columnwidth\else\width\fi}{!}{%
\begin{tabular}{@{}lccc@{}}
\toprule
\textbf{Layer} & \textbf{r(CAM, Edges)} & \textbf{r(CAM, Fringes)} & \textbf{Ratio} \\
\midrule
enc3    & 0.011 & $-$0.010 & 1.13 \\

enc4    & 0.242 &    0.183 & 1.33 \\

bottleneck & 0.144 & 0.130 & 1.10 \\

dec1    & 0.068 &    0.071 & 0.96 \\

dec3    & $-$0.007 & 0.007 & $-$0.72 \\

dec4    & 0.361 &    0.261 & 1.38 \\

\textbf{Average} & \textbf{0.136} & \textbf{0.107} & \textbf{1.28} \\
\bottomrule
\end{tabular}}
\end{table}

The three probes converge: the UNet baseline detects object boundaries and fills in learned shape templates rather than decoding phase from fringe intensity. Read against Section~\ref{sec:singleshot_theory}, the shortcut is the specific way an optimizer resolves the non-injectivity recapped there, by collapsing onto whatever shape-prior surrogate the training distribution supplies. The principled repair is therefore not more data or capacity but an architecture that routes reconstruction through wrapped phase and fringe order as in Eq.~\eqref{eq:phase_pipeline}, removing the shape-prior solution from the hypothesis space by construction. That is PhiCalNet.

\section{PhiCalNet: A Phase-Intermediate Architecture for Single-Shot FPP}
\label{sec:physics_aware}

The mechanistic analysis recapped in Section~\ref{sec:interpretability} showed that a depth-regressing UNet bypasses fringe physics and relies on object-shape priors, the same regime Section~\ref{sec:singleshot_theory} identifies as ill-posed under single-shot acquisition. The response those sections point to is to route the reconstruction through wrapped phase and fringe order (Eq.~\eqref{eq:phase_pipeline}) rather than learning $I_t \mapsto Z$ directly. To remove the shape-prior solution from the hypothesis space rather than merely discourage it via a loss penalty, we introduce \textbf{PhiCalNet}, a single-shot reconstruction architecture that realizes that pipeline. Its operative design choice can be stated in one sentence: the network outputs a wrapped-phase representation $(\sin\phi, \cos\phi)$ rather than depth, and the only path from that output to the reconstructed depth runs through a fixed differentiable calibration layer that applies the FPP triangulation physics deterministically. Two architectural pieces are load-bearing. \emph{Phase representation}: the network's output space is wrapped phase, not depth, foreclosing the shape-prior solution because depth is never produced as a direct learned quantity. \emph{Fixed differentiable calibration}: a non-learnable layer $\mathcal{C}(\cdot)$ maps unwrapped phase to depth using the camera--projector geometry. The layer is differentiable, which makes depth-supervised loss terms definable on its output (used as one component of the composite training loss in Section~\ref{sec:phaseunet_loss}) and activates a gradient path from the depth output back through the calibration into the learnable backbone that supports natural joint-learning extensions of the design (joint calibration learning, heteroscedastic depth likelihoods, multi-resolution depth losses, absolute-scale depth supervision; see Sections~\ref{sec:phaseunet_loss_ablation} and~\ref{sec:conclusion}); its primary architectural role for the headline configuration, however, is to convert phase to depth deterministically in the forward pass. Neither piece is unprecedented in isolation: phase-intermediate networks with downstream calibration math have been used in FPP before~\citep{feng2019fringe,nguyen2023fringe}, and depth-output networks are standard. Combining them, by making the network's output space wrapped phase and inserting a fixed differentiable calibration layer that converts that output to depth, is the architectural choice that removes the shape-prior solution from the hypothesis space without a loss penalty. The only path from the network's output to depth runs through $\mathcal{C}$, and the only way to produce accurate depth at the output of $\mathcal{C}$ is for the network's output to actually represent wrapped phase. The physics is therefore a hard architectural constraint, not a soft training signal. We describe the architecture in Section~\ref{sec:phaseunet_arch}, the training loss and a component-wise ablation in Section~\ref{sec:phaseunet_loss}, and report results in Section~\ref{sec:phaseunet_results}. Section~\ref{sec:pinn_contrast} runs the positive test of this design hypothesis: if the architectural framing is what is doing the work, then enforcing the same physics as a soft loss penalty on an otherwise unconstrained depth-regressing network (a PINN baseline) should \emph{not} match PhiCalNet, because shape-prior solutions remain in the hypothesis space. Section~\ref{sec:pinn_contrast} confirms exactly this.

\subsection{Architecture}
\label{sec:phaseunet_arch}

Let $\phi(u,v) \in [-\pi, \pi)$ denote the wrapped phase at pixel $(u,v)$ and $k(u,v) \in \{0, \ldots, K_{\max}-1\}$ its fringe order, so that the unwrapped phase is $\Phi(u,v) = \phi(u,v) + 2\pi\,k(u,v)$. For the horizontal fringe image we train on (frame 0), phase varies along the vertical axis of the $912 \times 1140$ projector pattern (1140 pixels) with period $T_p = 36$ pixels. As a result, the scene admits at most $\lfloor 1140/36 \rfloor = 31$ distinct fringe bands, and observed $k$ values span $[0, 31]$ across the test split. We therefore set $K_{\max} = 32$ to conservatively cover the observed range and clip spurious outliers left after gray-code decoding.

\begin{figure*}[pos=tp]
\centering
\includegraphics[width=\linewidth]{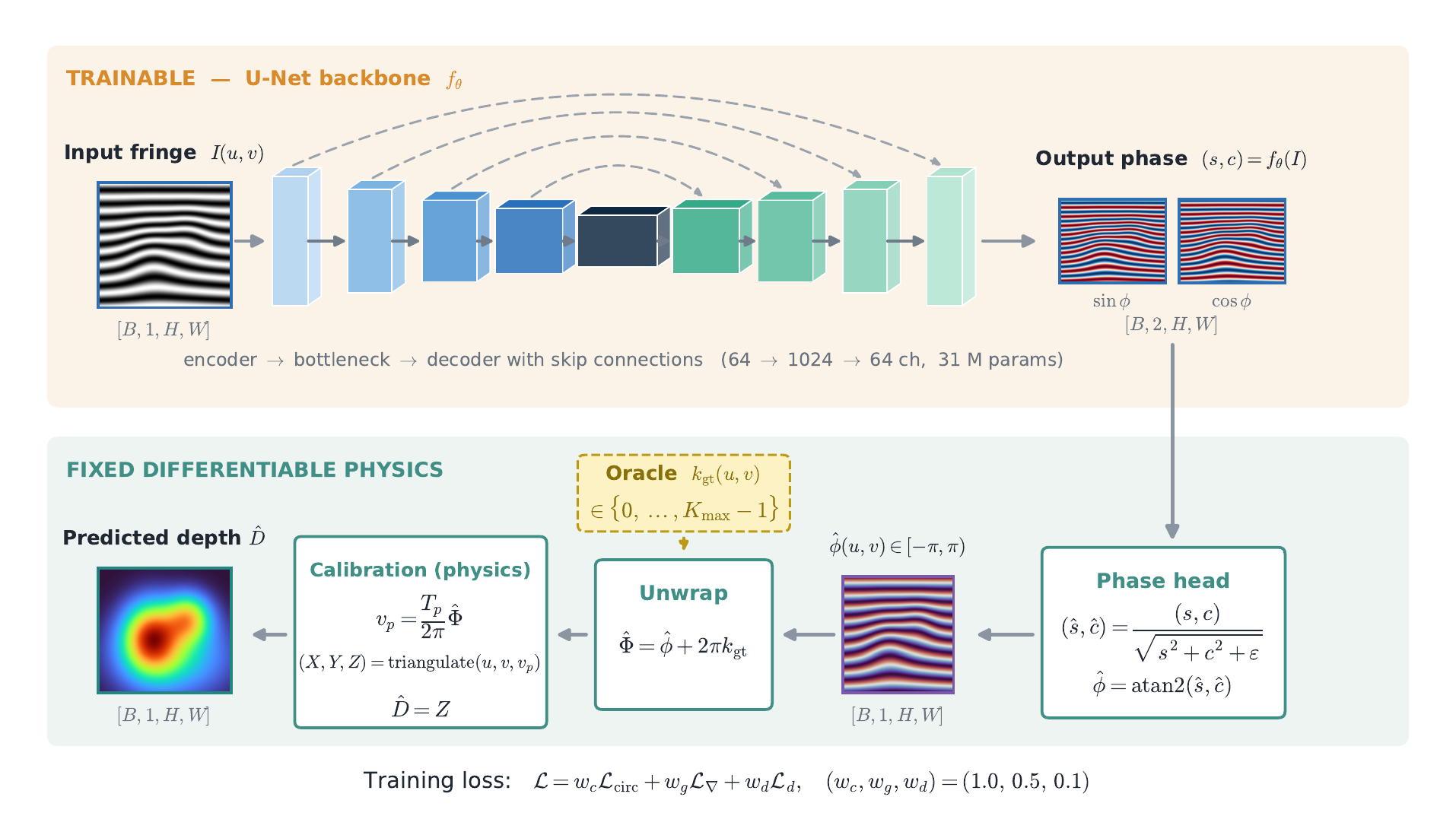}
\caption{PhiCalNet architecture. \textbf{Top row (learning)}: the input fringe $I$ is fed to a trainable UNet backbone (31~M parameters; encoder--bottleneck--decoder with a final $1{\times}1$ convolution) producing a two-channel field $(s, c) = (\sin\phi, \cos\phi)$. \textbf{Bottom row (physics)}: a fixed phase head maps $(s,c)$ to the wrapped phase $\hat{\phi}$ via $L_2$ unit-circle projection and $\mathrm{atan2}$; $\hat{\phi}$ and the auxiliary oracle fringe order $k_{\mathrm{gt}}$ then feed a fixed unwrap step that yields $\hat{\Phi} = \hat{\phi} + 2\pi k_{\mathrm{gt}}$, and a fixed calibration layer converts phase to a virtual projector column $v_p = T_p \hat{\Phi}/(2\pi)$ and triangulates with the camera--projector calibration to obtain $(X, Y, Z)$; the depth $\hat{D} = Z$. Only the UNet backbone has learnable parameters; all other operations are fixed and differentiable. Fringe, phase, and depth thumbnails are illustrative.}
\label{fig:phaseunet_architecture}
\end{figure*}

PhiCalNet (Fig.~\ref{fig:phaseunet_architecture}) composes four operations in series: a trainable UNet backbone $f_\theta$ that predicts $(s, c) = (\sin\phi, \cos\phi)$ from the fringe image $I$; a pixelwise unit-circle projection $(s,c) \mapsto (s,c)/\sqrt{s^2+c^2+\varepsilon}$ that enforces $\sin^2\phi + \cos^2\phi = 1$ and yields the wrapped phase $\hat{\phi} = \mathrm{atan2}(s,c)$; an oracle unwrap step that combines $\hat{\phi}$ with the ground-truth fringe order $k_{\mathrm{gt}}$ to produce $\hat{\Phi} = \hat{\phi} + 2\pi\,k_{\mathrm{gt}}$; and a fixed, non-learnable differentiable calibration layer $\mathcal{C}(\cdot)$ that maps the unwrapped phase to depth via the intrinsic and extrinsic calibration of the virtual projector--camera pair, $\hat{D} = \mathcal{C}(\hat{\Phi})$. The backbone shares its architecture with the baseline of Section~\ref{sec:baseline} (four encoder-decoder stages, 31~M parameters, instance normalization), and gradients flow through $\mathcal{C}$ back to all earlier layers, so any supervision applied to $\hat{D}$ propagates into the parameters that produce $\hat{\phi}$. The oracle-$k$ assumption isolates phase prediction from the separate, much harder problem of single-shot fringe-order classification, and is treated as an upper bound on what a phase-intermediate pipeline can deliver from one frame when fringe order is known from auxiliary information (in FPP-ML-Bench, $k_{\mathrm{gt}}$ is decoded from a 7-bit gray-code projection accompanying each fringe frame and median-filtered before use; in practice the same signal can come from a separate gray-code projection, a wide-period fringe in a prior frame, or a multi-frequency sequence).

Because the only trainable parameters are those of $f_\theta$, and the only path from $f_\theta$ to depth passes through $\mathcal{C}$, the representation produced at the output of $f_\theta$ must approximate wrapped phase for the depth to be accurate. The network is therefore physics-aware by construction.

\subsection{Loss Formulation}
\label{sec:phaseunet_loss}

PhiCalNet is trained with a composite loss whose components respect the geometry of the prediction target:
\begin{equation}
    \mathcal{L} = w_{c}\,\mathcal{L}_{\mathrm{circ}}
    + w_{g}\,\mathcal{L}_{\nabla}
    + w_{d}\,\mathcal{L}_{d},
\end{equation}
with weights $(w_c, w_g, w_d) = (1.0, 0.5, 0.1)$. The primary supervision is the geodesic phase term $\mathcal{L}_{\mathrm{circ}}$, a boundary-weighted distance on the unit circle,
\begin{equation}
    \mathcal{L}_{\mathrm{circ}} =
    \frac{\sum_{(u,v)\in\mathcal{M}} \alpha(u,v)\,\min\!\bigl(|\Delta\phi|,\,2\pi - |\Delta\phi|\bigr)}
         {\sum_{(u,v)\in\mathcal{M}} \alpha(u,v)},
\end{equation}
with $\Delta\phi = \hat{\phi}(u,v) - \phi_{\mathrm{gt}}(u,v)$, object mask $\mathcal{M}$, and boundary weight $\alpha(u,v) = 1 + \beta\,(|\phi_{\mathrm{gt}}|/\pi)^2$ ($\beta=3$) that ramps smoothly from $1$ in the fringe interior to $4$ at the $\pm\pi$ wrap boundary, up-weighting the regime that dominates classical spatial-unwrapping error~\citep{goldstein1988satellite, zuo2016temporal}; by measuring true angular separation it avoids the underestimation of wrap-boundary error incurred by an $L_1$ distance on $(\sin\phi, \cos\phi)$. The gradient term $\mathcal{L}_{\nabla}$ is masked $L_1$ on the spatial gradient of $(s,c)$, regularizing the predicted phase to be locally smooth except at the physical $2\pi$ jumps. The depth term $\mathcal{L}_{d}$ is the Hybrid $L_1$ ($\alpha=0.7$) on per-sample normalized depth, the same supervised loss identified as best for the baseline in Section~\ref{sec:baseline}, combining a masked object term with a global regularizer; because it is computed on $\hat{D} = \mathcal{C}(\hat{\Phi})$, its gradients propagate through the calibration layer $\mathcal{C}$. The training procedure is given in Algorithm~\ref{alg:phaseunet}; a component-wise ablation isolating the empirical contribution of each term is reported in Section~\ref{sec:phaseunet_loss_ablation} below.

\begin{algorithm}[!t]
\caption{PhiCalNet Training}
\label{alg:phaseunet}
\footnotesize
\begin{algorithmic}[1]
\Require Training set $\mathcal{T} = \{(I, k_{\mathrm{gt}}, D_{\mathrm{gt}}, \phi_{\mathrm{gt}}, \mathcal{M})\}$; calibration $\mathcal{C}$; weights $(w_c, w_g, w_d)$; boundary parameter $\beta$; learning rate $\eta$; small constant $\varepsilon$
\Ensure Trained backbone parameters $\theta$
\State Initialize $\theta$ using the same scheme as the baseline UNet
\For{each epoch}
  \For{each minibatch $(I, k_{\mathrm{gt}}, D_{\mathrm{gt}}, \phi_{\mathrm{gt}}, \mathcal{M}) \in \mathcal{T}$}
    \State $(s, c) \gets f_\theta(I)$ \Comment{two-channel CNN forward pass}
    \State $(s, c) \gets (s, c)/\sqrt{s^2 + c^2 + \varepsilon}$ \Comment{unit-circle projection}
    \State $\hat{\phi} \gets \mathrm{atan2}(s, c)$ \Comment{wrapped phase in $[-\pi, \pi)$}
    \State $\hat{\Phi} \gets \hat{\phi} + 2\pi\,k_{\mathrm{gt}}$ \Comment{unwrap with oracle order}
    \State $\hat{D} \gets \mathcal{C}(\hat{\Phi})$ \Comment{differentiable calibration to depth}
    \Statex \hspace{\algorithmicindent}\textit{Compute composite loss:}
    \State $\alpha(u,v) \gets 1 + \beta\,(|\phi_{\mathrm{gt}}(u,v)|/\pi)^2$
    \State $\mathcal{L}_{\mathrm{circ}} \gets \bigl[\textstyle\sum_{(u,v)\in\mathcal{M}} \alpha(u,v)\,\min(|\Delta\phi|,\,2\pi-|\Delta\phi|)\bigr] \big/ \bigl[\textstyle\sum_{(u,v)\in\mathcal{M}} \alpha(u,v)\bigr]$
    \State $\mathcal{L}_{\nabla} \gets \mathrm{MaskedL1}\bigl(\nabla(s,c),\,\nabla(\sin\phi_{\mathrm{gt}}, \cos\phi_{\mathrm{gt}});\,\mathcal{M}\bigr)$
    \State $\mathcal{L}_{d} \gets \mathrm{HybridL1}(\hat{D},\,D_{\mathrm{gt}};\,\mathcal{M})$ \Comment{on per-sample normalized depth, $\alpha=0.7$}
    \State $\mathcal{L} \gets w_c\,\mathcal{L}_{\mathrm{circ}} + w_g\,\mathcal{L}_{\nabla} + w_d\,\mathcal{L}_{d}$
    \Statex \hspace{\algorithmicindent}\textit{Backpropagation through $\mathcal{C}$ and $f_\theta$:}
    \State $\theta \gets \theta - \eta\,\nabla_\theta \mathcal{L}$
  \EndFor
\EndFor
\State \Return $\theta$
\end{algorithmic}
\end{algorithm}

\subsubsection{Component Ablation}
\label{sec:phaseunet_loss_ablation}

To establish what each loss term contributes empirically, we retrain three variants on the same data, optimizer, and schedule used for the reference PhiCalNet, each with one loss weight set to zero. Object-level and pixel-level metrics for the three variants and the full-loss reference are reported in Table~\ref{tab:phicalnet_loss_ablation}; all metrics are computed with the canonical raw-depth pipeline used in Tables~\ref{tab:phaseunet_results} and~\ref{tab:phaseunet_error_distribution}.

\begin{table}[pos=tbp]
\caption{PhiCalNet component ablation. Each row sets one loss weight to zero with all other training settings held fixed at the reference values; the reference row is the model used for all downstream analyses (Sections~\ref{sec:phaseunet_results}--\ref{sec:phicalnet_uq}). Object-level MAE and RMSE reported in millimeters. Pixel-wise metrics aggregated across the 30 test samples and 2{,}002{,}360 object pixels. The ``Tail'' column counts object pixels with absolute error above 100~mm.}
\label{tab:phicalnet_loss_ablation}

\centering
\small
\resizebox{\ifdim\width>\columnwidth\columnwidth\else\width\fi}{!}{%
\begin{tabular}{@{}lcccccccccc@{}}
\toprule
 & \multicolumn{2}{c}{\textbf{Object (mm)}} & \multicolumn{2}{c}{\textbf{Pixel (mm)}} & \multicolumn{4}{c}{\textbf{Pixel percentiles (mm)}} & \multicolumn{2}{c}{\textbf{Tail (err $>$ 100~mm)}} \\
\cmidrule(lr){2-3}\cmidrule(lr){4-5}\cmidrule(lr){6-9}\cmidrule(lr){10-11}
\textbf{Variant} & MAE & RMSE & MAE & RMSE & P50 & P90 & P99 & P99.9 & Pixels & \% \\
\midrule
Full loss (reference)                                  & 4.46          & 17.82          & 4.07          & 20.28          & 1.79          & 8.92          & 27.14          & 103.1          & 2{,}060          & 0.103 \\

$w_c = 0$  (drop $\mathcal{L}_{\mathrm{circ}}$)        & 6.03          & 25.17          & 5.68          & 27.66          & 2.75          & 10.92         & 31.94          & 282.9          & 4{,}761          & 0.238 \\

$w_g = 0$  (drop $\mathcal{L}_{\nabla}$)               & 4.71          & 18.20          & 4.19          & 20.83          & 1.89          & 9.11          & 26.23          & 102.4          & 2{,}055          & 0.103 \\

$w_d = 0$  (drop $\mathcal{L}_{d}$)                    & \textbf{4.27} & \textbf{16.41} & \textbf{3.76} & \textbf{19.37} & \textbf{1.63} & \textbf{8.24} & \textbf{25.09} & \textbf{84.0}  & \textbf{1{,}621} & \textbf{0.081} \\
\bottomrule
\end{tabular}}
\end{table}

The wrap-aware geodesic term $\mathcal{L}_{\mathrm{circ}}$ is the primary driver. Removing it raises object MAE by 35\% ($4.46 \to 6.03$~mm), raises object RMSE by 41\% ($17.82 \to 25.17$~mm), and more than doubles the population of object pixels with absolute error above 100~mm ($2{,}060 \to 4{,}761$, or $0.103\% \to 0.238\%$ of the object mask), with the 99.9th-percentile pixel error rising from 103 to 283~mm. The boundary-weighted angular geometry of the loss is therefore load-bearing, consistent with the wrap-boundary localization of the residual error analyzed in Section~\ref{sec:phaseunet_error_analysis}.

The smoothness term $\mathcal{L}_{\nabla}$ contributes a small but real bulk-distribution improvement: removing it raises object MAE by 6\% (4.71 vs.\ 4.46~mm) and pixel-wise MAE by 3\% (4.19 vs.\ 4.07~mm), with the heavy-tail percentiles and wrap-boundary tail count unchanged. It is retained as a standard smoothness regularizer that mildly tightens the bulk of the error distribution without affecting the wrap-boundary failure mode.

Removing the depth term $\mathcal{L}_{d}$ uniformly improves every reported metric for the headline configuration: object MAE drops from 4.46 to 4.27~mm, object RMSE from 17.82 to 16.41~mm, pixel-wise RMSE from 20.28 to 19.37~mm, and the wrap-boundary pixel tail from 2{,}060 to 1{,}621 pixels (a 21\% reduction), with the 99.9th-percentile pixel error dropping from 103 to 84~mm. This matches what the architecture predicts: with phase as the network's output and depth a deterministic function of phase through the fixed calibration layer (Section~\ref{sec:phaseunet_arch}), accurate phase already implies accurate depth in the single-shot phase-only configuration, so a supervised depth term on per-sample normalized depth adds marginal training signal beyond what $\mathcal{L}_{\mathrm{circ}}$ already supplies and slightly competes with the optimum reached by $\mathcal{L}_{\mathrm{circ}}$ alone.

$\mathcal{L}_{d}$ is nevertheless retained at $w_d = 0.1$ in the reported model because it activates a depth-supervised gradient path that runs from the depth output back through the calibration layer $\mathcal{C}$ and into every learnable parameter of the UNet backbone. That gradient path is the architectural feature that distinguishes a \emph{fixed differentiable} calibration from a fixed non-differentiable one, and it is the path on which the natural extensions of the PhiCalNet design are built: joint learning of calibration intrinsics or extrinsics, heteroscedastic depth likelihoods that allow per-pixel noise modeling, multi-resolution or perceptual depth losses, and depth supervision on the absolute rather than per-sample-normalized scale all require gradients to flow from the depth output back through $\mathcal{C}$ and into the learnable backbone. Keeping $\mathcal{L}_{d}$ active in the reported checkpoint exercises that gradient path during training and gives the downstream MI, UQ, multi-frame, and sensitivity analyses (Sections~\ref{sec:phaseunet_results}--\ref{sec:phicalnet_uq}) a faithful baseline against which those follow-up variants can be evaluated. The decomposition is then straightforward: $\mathcal{L}_{\mathrm{circ}}$ is the operative training signal for the headline single-shot phase-only output, $\mathcal{L}_{\nabla}$ tightens the bulk of the error distribution as a mild smoothness regularizer, and $\mathcal{L}_{d}$ is the gradient-path infrastructure on which the joint-learning extensions of PhiCalNet will build, with the 4\% MAE gap between $w_d = 0.1$ and $w_d = 0$ leaving every qualitative conclusion in the downstream sections unchanged.

\subsection{Results}
\label{sec:phaseunet_results}

With each loss term's contribution isolated, we now compare the reported PhiCalNet model, trained with the full composite loss, to the depth-regressing UNet baseline of Section~\ref{sec:baseline}. Table~\ref{tab:phaseunet_results} reports the comparison on the 30-sample test split. PhiCalNet reduces object MAE from 14.54~mm to 4.46~mm, a 3.3$\times$ improvement, while object RMSE remains near baseline (17.88~mm vs.\ 17.82~mm). The disparity between these two metrics is itself informative and we analyze it in Section~\ref{sec:phaseunet_error_analysis}.

\begin{table}[pos=tbp]
\caption{Single-shot single-frame comparison on the 30-sample test split. PhiCalNet operates under the oracle fringe-order assumption described in Section~\ref{sec:phaseunet_arch}. Object-level MAE and RMSE reported in millimeters.}
\label{tab:phaseunet_results}

\centering
\small
\resizebox{\ifdim\width>\columnwidth\columnwidth\else\width\fi}{!}{%
\begin{tabular}{@{}lcc@{}}
\toprule

\textbf{Model} & \textbf{MAE} & \textbf{RMSE} \\
\midrule
UNet baseline (Hybrid L1)         & 14.54 & 17.88 \\

\textbf{PhiCalNet}                & \textbf{4.46} & \textbf{17.82} \\
\bottomrule
\end{tabular}}
\end{table}

\subsection{Positive Test of the Design Hypothesis: Loss-Level Physics Does Not Suffice}
\label{sec:pinn_contrast}

The 3.3$\times$ MAE improvement reported above follows from moving physics into the forward pass. The design hypothesis of Section~\ref{sec:physics_aware} claims more sharply that this architectural framing is what is doing the work, and not merely the presence of calibration physics anywhere in the pipeline. The positive test of that hypothesis is the obvious alternative: keep depth as the network output but enforce the same physics as a soft loss penalty on the otherwise unconstrained depth-regressing network, which is the standard physics-informed neural network approach the machine learning literature would suggest by default.~\citep{raissi2019pinn,karniadakis2021physics} If the soft-penalty version closes the gap, the architectural framing is overdetermined; if it does not, the architectural choice is isolated as the operative factor. We refer to this alternative as PINN-UNet: the baseline UNet that predicts normalized depth $\hat{D}_{\mathrm{norm}}$, augmented with a differentiable inverse-calibration layer $\mathcal{D}(\cdot)$ that maps predicted depth back to the phase that should have produced it, and trained with the objective
\begin{equation}
    \mathcal{L}_{\mathrm{PINN}} =
    \mathcal{L}_{\mathrm{HybridL1}}
    + \lambda \cdot \bigl\|\mathcal{D}(\hat{D}) - \Phi_{\mathrm{gt}}\bigr\|_{1,\mathcal{M}},
    \label{eq:pinn_loss}
\end{equation}
where $\mathcal{L}_{\mathrm{HybridL1}}$ is the best-performing supervised loss from Section~\ref{sec:baseline} and $\lambda$ controls the strength of the physics term. The intent of Eq.~(\ref{eq:pinn_loss}) is that minimizing $\mathcal{L}_{\mathrm{PINN}}$ requires the network to produce depths that are both close to the ground truth and physically consistent with the observed fringe. Crucially, PINN-UNet and PhiCalNet are built on the same camera--projector calibration; only the point at which the physics is enforced (soft loss term vs.\ forward pass) differs.

We trained four PINN-UNet variants at $\lambda \in \{0.005, 0.01, 0.02, 0.05\}$ using the same data, optimizer, and schedule as the baseline UNet in Section~\ref{sec:baseline}. Table~\ref{tab:pinn_results} reports the results.

\begin{table}[pos=tbp]
\caption{PINN-UNet variants on the 30-sample test split, compared to the baseline depth-regressing UNet and to PhiCalNet. Object-level MAE and RMSE reported in millimeters.}
\label{tab:pinn_results}

\centering
\small
\resizebox{\ifdim\width>\columnwidth\columnwidth\else\width\fi}{!}{%
\begin{tabular}{@{}lcc@{}}
\toprule

\textbf{Model} & \textbf{MAE} & \textbf{RMSE} \\
\midrule
UNet baseline (Hybrid L1)    & 14.54 & 17.88 \\

PINN-UNet, $\lambda=0.005$   & 16.44 & 20.42 \\

PINN-UNet, $\lambda=0.01$    & 14.87 & 18.63 \\

PINN-UNet, $\lambda=0.02$    & 16.50 & 21.26 \\

PINN-UNet, $\lambda=0.05$    & 17.05 & 21.87 \\

\textbf{PhiCalNet (ours)}    & \textbf{4.46} & \textbf{17.82} \\
\bottomrule
\end{tabular}}
\end{table}

Across the $\lambda$ sweep the physics term produces no measurable gain over the baseline. The best variant ($\lambda=0.01$) is statistically indistinguishable from the baseline (14.87 vs.\ 14.54~mm object MAE), and larger $\lambda$ values degrade performance because the physics gradient competes with the supervised signal without constraining the representation. This pattern is consistent with a documented failure mode of physics-informed training. Krishnapriyan et al.~\citep{krishnapriyan2021characterizing} show that PINN losses do not, on their own, compel the network to learn the intended physical quantity when a lower-loss shortcut is available to the optimizer. Karniadakis et al.~\citep{karniadakis2021physics} and subsequent work~\citep{wang2022respecting} have documented that the balance between supervised and physics terms is unstable and representation-dependent, and that the physics residual can be driven to low values by solutions that do not correspond to the target physics. In our setting the consequence is direct: as long as the network predicts depth, the physics penalty measures the residual of an already-learned depth field against a reconstruction that depends on it, and a shape-correct depth map produced through boundary detection can satisfy the physics term approximately without the network ever representing phase internally. The mechanistic result recapped in Section~\ref{sec:interpretability} therefore persists even under a physics-informed loss. Moving physics into the architecture, as PhiCalNet does, is what forces phase to be used and produces the 3.3$\times$ improvement reported in Table~\ref{tab:phaseunet_results}.

\section{Error Distribution and Residual Failure Mode}
\label{sec:phaseunet_error_analysis}

The disparity between the PhiCalNet object MAE (4.46~mm) and object RMSE (17.82~mm) is unusually large and warrants a pixel-level analysis. Aggregating absolute errors across all 2{,}002{,}360 object pixels in the test split, the pixel-wise MAE is 4.07~mm and the pixel-wise RMSE is 20.28~mm. The median error is 1.79~mm, the 90th percentile is 8.9~mm, the 99th is 27.1~mm, and the 99.9th is 103~mm; the maximum reaches 1515~mm. Table~\ref{tab:phaseunet_error_distribution} summarizes the key percentiles and threshold counts.

\begin{table}[pos=tbp]
\caption{Per-pixel error percentiles and tail counts for PhiCalNet (30 test samples, 2{,}002{,}360 object pixels). The 0.103\% of pixels above 100~mm carry essentially all of the RMSE.}
\label{tab:phaseunet_error_distribution}

\centering
\small
\resizebox{\ifdim\width>\columnwidth\columnwidth\else\width\fi}{!}{%
\begin{tabular}{@{}lclc@{}}
\toprule
\textbf{Percentile} & \textbf{Value (mm)} & \textbf{Threshold} & \textbf{Pixels (\%)} \\
\midrule
50th    & 1.79    &  err $>$ 10~mm    & 167{,}176 (8.35\%) \\

75th    & 4.00    &  err $>$ 50~mm    & 4{,}414  (0.22\%) \\

90th    & 8.92    &  err $>$ 100~mm   & 2{,}060  (0.103\%) \\

99th    & 27.14   &  err $>$ 500~mm   & 1{,}059  (0.053\%) \\

99.9th  & 103.05  &  Max err          & 1515.14~mm \\

99.99th & 986.39  &  Pixel-wise MAE / RMSE  & 4.07 / 20.28 \\
\bottomrule
\end{tabular}}
\end{table}

The distribution is heavy-tailed. Half of all object pixels have absolute error below 1.79~mm, and 90\% are below 8.92~mm, which is more than an order of magnitude better than the baseline UNet. The entirety of the 20~mm RMSE is carried by the top 0.1\% of pixels: 2{,}060 pixels (0.103\% of the object mask) exceed 100~mm error, and 1{,}059 (0.053\%) exceed 500~mm, with individual pixels reaching 1515~mm.

\begin{figure*}[pos=tp]
    \centering
    \begin{subfigure}[t]{\linewidth}
        \centering
        \includegraphics[width=\linewidth]{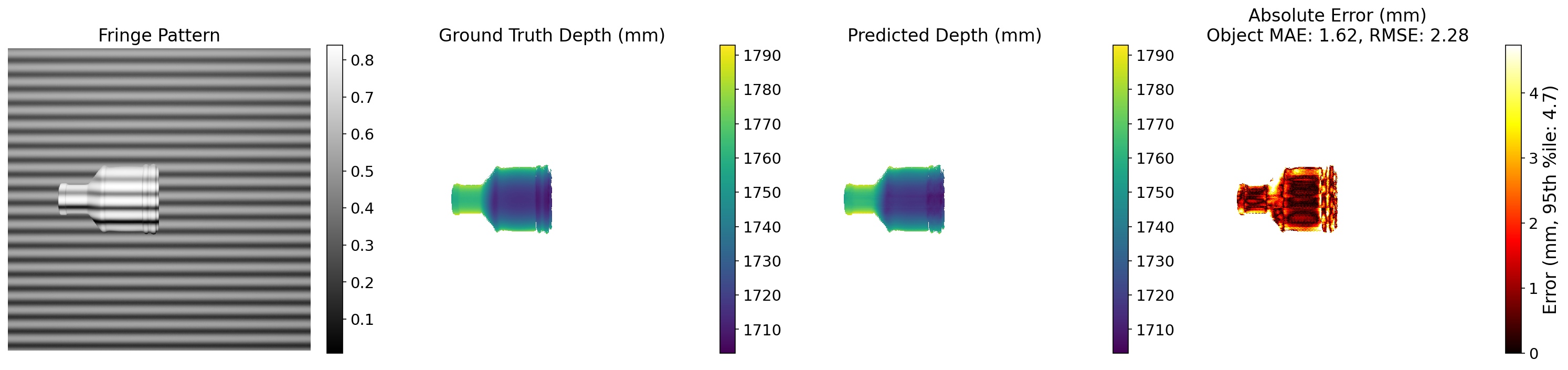}
        \caption{Depth set: fringe, ground-truth depth, predicted depth, absolute depth error. The error map is diffuse and low throughout.}
    \end{subfigure}
    \\[0.4em]
    \begin{subfigure}[t]{\linewidth}
        \centering
        \includegraphics[width=\linewidth]{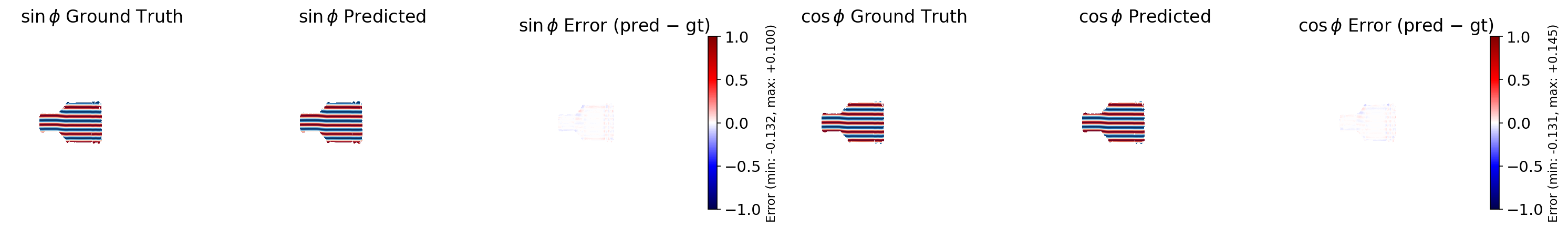}
        \caption{Wrapped-phase components with per-component signed error maps. The sparse, low-amplitude error bands are consistent with the object only spanning a few fringe periods.}
    \end{subfigure}
    \caption{PhiCalNet predictions on the clean test sample \texttt{container\_bottle\_A180} (object MAE 1.63~mm, RMSE 2.28~mm). Shallow depth variation keeps the object within a few fringe bands. The depth error panel is clipped at the 95th object percentile for visibility; the $\sin\phi$ and $\cos\phi$ error panels use a fixed symmetric range in $[-1, +1]$ so the near-zero bulk appears near-white and any wrap-boundary sign flips stand out in red or blue.}
    \label{fig:phaseunet_clean_sample}
\end{figure*}

\begin{figure*}[pos=tp]
    \centering
    \begin{subfigure}[t]{\linewidth}
        \centering
        \includegraphics[width=\linewidth]{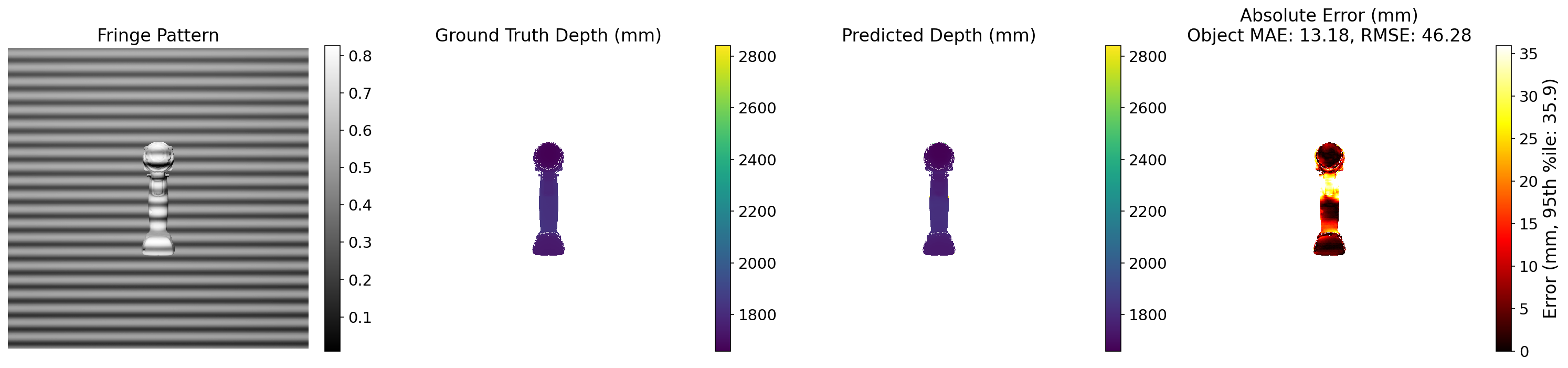}
        \caption{Depth set. Error concentrates in horizontal stripes that trace the $\pm\pi$ wrap lines of the true wrapped phase (visible in the depth GT as the iso-contours between fringe bands).}
    \end{subfigure}
    \\[0.4em]
    \begin{subfigure}[t]{\linewidth}
        \centering
        \includegraphics[width=\linewidth]{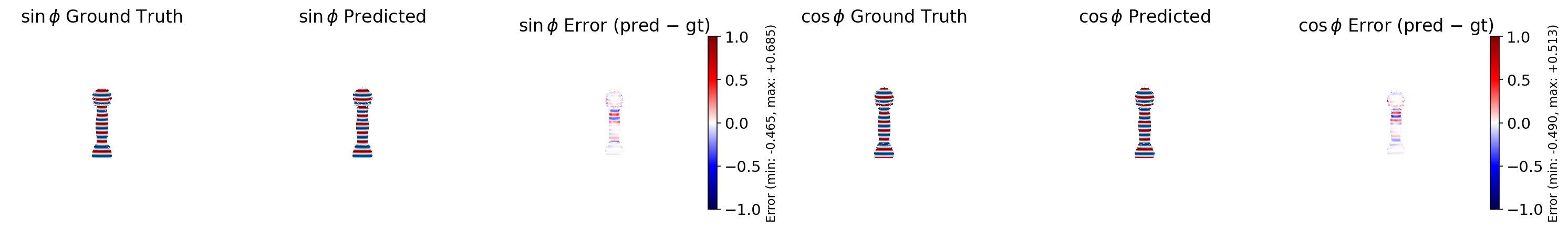}
        \caption{Wrapped-phase components with per-component signed error maps. From left to right in each pair: ground truth, predicted, and error (pred~$-$~gt) for $\sin\phi$ and $\cos\phi$. GT and predicted maps are visually almost indistinguishable, but the error maps reveal sharp red/blue bands aligned exactly with the boundaries between consecutive fringe periods, i.e., the locus where $\phi_{\mathrm{gt}}$ crosses $\pm\pi$. A small Euclidean error in $(s,c)$ across the wrap translates into a $\pm 2\pi$ jump in $\mathrm{atan2}$, producing one fringe period of depth error in (a).}
    \end{subfigure}
    \caption{PhiCalNet predictions on the failure test sample \texttt{power\_drill\_A0} (object MAE 13.18~mm, RMSE 46.28~mm). Complex multi-level geometry makes the object surface cross many $\pm\pi$ wrap lines, and the resulting wrap-boundary sign errors accumulate into the horizontal stripes seen in the depth error map. Colorbar conventions follow Fig.~\ref{fig:phaseunet_clean_sample}.}
    \label{fig:phaseunet_failure_sample}
\end{figure*}

To understand what these outliers correspond to, recall that the fringe order $k$ is supplied by oracle in the PhiCalNet forward pass (Section~\ref{sec:phaseunet_arch}), so the integer part of the unwrapped phase is always correct. The errors therefore originate entirely in the predicted wrapped phase $\hat{\phi}$. The relevant failure mode is not a misclassified $k$ but a \emph{wrap-boundary sign error}: at pixels where the ground-truth wrapped phase lies close to the $\pm\pi$ discontinuity, the network's predicted $(\hat{s},\hat{c})$ can land just on the wrong side of the wrap. The Euclidean error in $(\sin\phi,\cos\phi)$ space is small in this case (which is why the predicted and ground-truth wrapped-phase images in Figs.~\ref{fig:phaseunet_clean_sample} and \ref{fig:phaseunet_failure_sample} look visually identical), but $\mathrm{atan2}(\hat{s},\hat{c})$ returns a value on the opposite side of the wrap, so $\hat{\phi}$ differs from $\phi_{\mathrm{gt}}$ by approximately $\pm 2\pi$. When the oracle $k$ is added to this wrongly-signed wrapped phase, the unwrapped phase $\hat{\Phi}$ is displaced by $\pm 2\pi$, which after calibration corresponds to approximately one fringe period of depth error (50--100~mm in our geometry), with multi-wrap excursions producing the observed extreme values.

Figures~\ref{fig:phaseunet_clean_sample} and~\ref{fig:phaseunet_failure_sample} localize these errors spatially. The clean case (Fig.~\ref{fig:phaseunet_clean_sample}) has object RMSE 2.3~mm; the object has shallow depth variation, only a few $\pm\pi$ wrap lines cross its surface, and the error map is diffuse and low throughout. The failure case (Fig.~\ref{fig:phaseunet_failure_sample}a) has object RMSE 46.3~mm; the object has complex multi-level geometry and its surface depth crosses many $\pm\pi$ wrap lines, and the depth error map is organized into \emph{horizontal stripes}. The $\sin\phi$ and $\cos\phi$ error panels in Fig.~\ref{fig:phaseunet_failure_sample}(b) make the mechanism explicit: the error signal is concentrated in sharp red/blue bands that coincide with the boundaries between consecutive fringe periods in the ground-truth $(\sin\phi, \cos\phi)$ maps, i.e., the locus of pixels where $\phi_{\mathrm{gt}}$ crosses $\pm\pi$. The bands are not distributed noise; they are the geometric signature of residual wrap-boundary sign errors at the specific depth contours where the true wrapped phase approaches the wrap. The same pattern is visible in all high-RMSE samples.

\subsection{Oracle Outlier Exclusion}
\label{sec:oracle_exclusion}

To quantify how concentrated the error tail is, we evaluate an oracle-exclusion bound on the same predictions. For each threshold $T \in \{100, 200, 500\}$~mm we remove pixels with absolute error exceeding $T$ and recompute MAE and RMSE over the remaining pixels. This is an upper bound on any post-hoc correction scheme: a real method must identify the bad pixels without access to ground truth, and the oracle identifies them exactly.

\begin{table}[pos=tbp]
\caption{Oracle-exclusion analysis on PhiCalNet predictions. Pixels with absolute error greater than the threshold are removed from the test object mask before MAE and RMSE are recomputed. The ceiling shows what any post-hoc outlier rejector could achieve at best. Object-level MAE and RMSE reported in millimeters.}
\label{tab:oracle_exclusion}

\centering
\small
\resizebox{\ifdim\width>\columnwidth\columnwidth\else\width\fi}{!}{%
\begin{tabular}{@{}lcccc@{}}
\toprule

\textbf{Condition} & \textbf{Pixels Removed} & \textbf{\% of Mask} & \textbf{MAE} & \textbf{RMSE} \\
\midrule
No removal               &       0 &  0.000\% & 4.07 & 20.28 \\

Exclude err $>$ 500~mm   &  1{,}059 &  0.053\% & 3.64 &  7.54 \\

Exclude err $>$ 200~mm   &  1{,}221 &  0.061\% & 3.62 &  7.04 \\

Exclude err $>$ 100~mm   &  2{,}060 &  0.103\% & 3.57 &  6.46 \\
\bottomrule
\end{tabular}}
\end{table}

Removing 0.103\% of the object mask reduces RMSE from 20.28~mm to 6.46~mm, a 3.1$\times$ reduction, while MAE moves only from 4.07 to 3.57~mm (Table~\ref{tab:oracle_exclusion}). This ceiling is informative in two ways. First, PhiCalNet is near-oracle on 99.9\% of pixels. Second, the remaining RMSE gap is not a fitting error that additional loss engineering is likely to close. The ground-truth wrapped phase is a discontinuous function at every $\pm\pi$ crossing; no continuous regressor can track it exactly there, and whenever the network's $(\hat{s},\hat{c})$ straddles the wrap, the resulting $\mathrm{atan2}$ output lies on the wrong side. The magnitude of that mistake is independent of how close the prediction is: even a vanishingly small sign error produces a full $2\pi$ displacement in $\hat{\phi}$, and therefore a full fringe period of depth error. Classical spatial phase-unwrapping algorithms encounter a structurally similar failure in the same regime.~\citep{goldstein1988satellite, zuo2016temporal} Multi-frame phase shifting resolves the ambiguity by construction because the additional frames reconstruct $\phi_{\mathrm{gt}}$ analytically per pixel, removing the need for a continuous regressor to approximate a discontinuous target. The 3.8$\times$ object-MAE gap between single-shot and multi-shot reconstruction observed throughout this paper therefore reflects an information-theoretic deficit of single-frame acquisition rather than a modeling limitation of the learner.

\section{Multi-Frame Extension and Fringe-Order Sensitivity}
\label{sec:phicalnet_multiframe}

The preceding error analysis argues on theoretical grounds that the single-shot residual is information-limited and that fringe-order information is necessary for single-shot reconstruction to be well-posed. We now validate both claims empirically. Three experiments are reported: (i) a multi-frame extension of PhiCalNet to verify that more observational information closes the wrap-boundary gap, (ii) a learned-$k$ ablation that removes the oracle assumption and observes how the pipeline degrades, and (iii) a fringe-order sensitivity sweep that quantifies how much upstream $k$ noise the pipeline can absorb without degrading depth accuracy.

\subsection{Multi-Frame Input Resolves the Single-Shot Residual}
\label{sec:phicalnet_multiframe_scaling}

We retrain PhiCalNet with the three-frame phase-shifted stack $\{F_0, F_6, F_{12}\}$ as input (channel-stacked $[B, 3, H, W]$), spaced exactly $120^\circ$ apart under the 18-step generator so they form a classical 3-step phase-shift set. All other factors (loss formulation, optimizer, oracle-$k$ assumption) are held fixed. Table~\ref{tab:phicalnet_multiframe_scaling} reports the result alongside the existing single-shot baseline.

\begin{table}[pos=tbp]
\caption{Frame-count scaling for PhiCalNet. All rows use oracle fringe order $k_{\mathrm{gt}}$. Object-level MAE and RMSE reported in millimeters.}
\label{tab:phicalnet_multiframe_scaling}

\centering
\small
\resizebox{\ifdim\width>\columnwidth\columnwidth\else\width\fi}{!}{%
\begin{tabular}{@{}lccc@{}}
\toprule
\textbf{Configuration} & \textbf{Frames} & \textbf{MAE} & \textbf{RMSE} \\
\midrule
PhiCalNet (single-shot) & 1 & 4.46 & 17.82 \\

\textbf{PhiCalNet (3-frame)} & \textbf{3} & \textbf{1.16} & \textbf{7.24} \\
\bottomrule
\end{tabular}}
\end{table}

Adding two phase-shifted frames drops object MAE from 4.46 to 1.16~mm (3.8$\times$ reduction) and collapses the wrap-boundary RMSE tail from 17.82 to 7.24~mm (2.5$\times$). At the pixel level, the fraction of object pixels with absolute error above 100~mm falls from 0.103\% (single-shot) to 0.011\% (3-frame), an order of magnitude reduction in the catastrophic tail. The 3-frame result closely tracks what the information-theoretic argument of Section~\ref{sec:phaseunet_error_analysis} predicts: with three phase-shifted observations, $\hat\phi$ is analytically recoverable per pixel, and the continuous regressor no longer has to approximate a discontinuous target. The residual $\sim 1$~mm MAE is the accumulated effect of noise in the rendered dataset plus the 3-step versus 18-step precision gap ($\sim 8^\circ$ phase on average).

\subsection{Learning Fringe Order from Single or Multi-Frame Input}
\label{sec:phicalnet_learnedk}

To test whether the oracle-$k$ assumption can be removed, we train three variants that predict fringe order $k$ from the network input rather than receiving it as ground truth. Let \textit{Variant L} denote fully-learned phase and fringe order (a PhiCalNet backbone augmented with a fringe-order classification head trained by cross-entropy against $k_{\mathrm{gt}}$), and \textit{Variant H} a hybrid architecture in which wrapped phase is computed analytically from the three input frames via the closed-form 3-step formula and only a learned $k$-head is trained, splitting learned and analytical components along the boundary where physics is ambiguous versus exact. Table~\ref{tab:phicalnet_learnedk} reports each at 1-frame and 3-frame input where applicable; Variant H requires $\geq 3$ frames and is only defined for the 3-frame case.

\begin{table}[pos=tbp]
\caption{Learned fringe-order variants on the 30-sample test split. All models use the same PhiCalNet calibration pipeline; only the $k$-source and phase-source differ. Variant L: learned $\hat\phi$ and learned $k$. Variant L$^+$: same as L with a wider $k$-head (256-channel, 5-block, 3.27~M head parameters vs.\ 0.38~M default, an 8.6$\times$ head-capacity increase). Variant H: analytical $\hat\phi$ and learned $k$ (hybrid). $k$ accuracy is the fraction of object pixels where $\arg\max \hat k = k_{\mathrm{gt}}$. Object-level MAE and RMSE reported in millimeters.}
\label{tab:phicalnet_learnedk}

\centering
\small
\resizebox{\ifdim\width>\columnwidth\columnwidth\else\width\fi}{!}{%
\begin{tabular}{@{}lccccc@{}}
\toprule
\textbf{Variant} & \textbf{Frames} & \textbf{Trainable} & \textbf{$k$ acc.} & \textbf{MAE} & \textbf{RMSE} \\
\midrule
L (learned $\hat\phi$, learned $k$)        & 1 & 31.4~M & 36.4\% &  799.7 & 1209.6 \\

L (learned $\hat\phi$, learned $k$)        & 3 & 31.4~M & 50.1\% &  595.8 & 1053.3 \\

L$^+$ (learned $\hat\phi$, wide $k$-head)  & 3 & 34.3~M & 39.5\% &  666.1 & 1041.0 \\

H (analytical $\hat\phi$, learned $k$)     & 3 &  0.4~M & 19.8\% & 1439.3 & 2031.4 \\
\bottomrule
\end{tabular}}
\end{table}

All three variants produce catastrophic depth errors ($\geq$596~mm object MAE). Per-sample analysis reveals the failure mode is not training instability but \emph{class-specific memorization}: across all three configurations, the per-pixel $k$ accuracy reaches $\geq$91\% on smooth-geometry samples (all six paint-container-spraycan views in the test split) and collapses to 11--35\% on complex multi-level geometry (wooden-boards, magazine-stack, power-drill views). The network absorbs the easy geometry via shape priors, the same pathology recapped in Section~\ref{sec:interpretability} (established by \citealp{haroon2026diagnosis}), and fails on geometries where shape priors do not disambiguate fringe order.

Counter-intuitively, the capacity-rich fully-learned Variant L at 3 frames outperforms the capacity-sparse hybrid Variant H at 3 frames (596 vs.\ 1439~mm MAE) despite the hybrid receiving physics-exact wrapped phase as input. Variant L$^+$, which adds 8.6$\times$ more parameters to the $k$-head specifically (3.27~M vs.\ 0.38~M), sharpens the picture further: it achieves a lower training validation loss (0.79 vs.\ 1.03) but a \emph{higher} test object MAE (666 vs.\ 596~mm) and \emph{lower} overall $k$ accuracy (39.5\% vs.\ 50.1\%). Per-sample analysis of Variant L$^+$ shows the regression is geometry-specific: on the smooth-geometry spraycan samples it reduces the best-case MAE (14--20~mm vs.\ 24--34~mm for Variant L) while on complex geometries it degrades further (worst-case 2153~mm vs.\ 1950~mm). The widened head is a \emph{better memorizer} of training-distribution geometries but a \emph{worse generalizer} to test-distribution geometries, extending the per-sample class-specific failure signature visible in both Variant L and Variant H.

These three variants, whose $k$-head capacity spans an 8.6$\times$ range (0.4~M / 0.4~M / 3.3~M dedicated to $k$, or 0.4~M / 31.4~M / 34.3~M including the phase backbone), all fail in the same structured way: near-perfect performance on smooth geometries where shape priors disambiguate the fringe order, and catastrophic failure on complex multi-level geometries where they do not. Additional capacity moves the model further into the memorization regime rather than toward a successful general $k$-predictor. We therefore conclude that $k$-prediction from a small number of fringe frames is an ill-posed problem under the information content available, rather than an under-trained or under-parameterized one.

\subsection{Sensitivity to Imperfect Fringe-Order Input}
\label{sec:phicalnet_noisyk}

The learned-$k$ ablation of Section~\ref{sec:phicalnet_learnedk} does not by itself quantify how close a $k$-source must be to perfect before the pipeline becomes unusable. To separate ``oracle $k$ is assumed'' from ``$k$ must be pixel-perfect'', we evaluate both 1-frame and 3-frame PhiCalNet on deliberately corrupted fringe-order inputs. Two corruption modes are tested: \emph{off-by-one}, in which a random fraction of object pixels has $k$ perturbed by $\pm 1$ (matching gray-code decode boundary noise), and \emph{uniform}, in which the same fraction is replaced by a uniformly random integer in $[0, K_{\max} - 1]$ (a worst-case decoder failure). Table~\ref{tab:phicalnet_noisyk} sweeps the corruption fraction from 0 to 25\% for each mode at both frame counts; model phase predictions are held fixed across corruption levels so $k$ noise is the only controlled variable.

\begin{table}[pos=tbp]
\caption{PhiCalNet with deliberately corrupted oracle fringe order. ``off-1'' corruption perturbs the selected pixels' $k$ by $\pm 1$ (gray-code-like boundary noise); ``uniform'' replaces them with a random integer in $[0, K_{\max}-1]$. Object-level MAE and RMSE reported in millimeters. Fraction 0 rows reproduce the clean oracle-$k$ numbers from Table~\ref{tab:phicalnet_multiframe_scaling}.}
\label{tab:phicalnet_noisyk}

\centering
\small
\resizebox{\ifdim\width>\columnwidth\columnwidth\else\width\fi}{!}{%
\begin{tabular}{@{}lccccc@{}}
\toprule
& & \multicolumn{2}{c}{\textbf{1-frame PhiCalNet}} & \multicolumn{2}{c}{\textbf{3-frame PhiCalNet}} \\
\cmidrule(lr){3-6}
\textbf{Mode} & \textbf{Fraction} & \textbf{MAE} & \textbf{RMSE} & \textbf{MAE} & \textbf{RMSE} \\
\midrule
(oracle)   & 0\%  & 4.46 & 17.82 & 1.16 &  7.24 \\

off-by-one & 1\%  & 4.47 & 18.10 & 1.17 &  7.87 \\

off-by-one & 5\%  & 4.56 & 20.09 & 1.25 & 11.11 \\

off-by-one & 10\% & 4.78 & 25.39 & 1.48 & 17.83 \\

off-by-one & 25\% & 6.75 & 49.14 & 3.37 & 44.21 \\

uniform    & 1\%  & 4.48 & 18.61 & 1.17 &  7.96 \\

uniform    & 5\%  & 4.65 & 24.32 & 1.29 & 13.28 \\

uniform    & 10\% & 5.14 & 35.38 & 1.54 & 21.18 \\

uniform    & 25\% & 8.67 & 79.60 & 5.75 & 76.80 \\
\bottomrule
\end{tabular}}
\end{table}

Both configurations show the same qualitative response to $k$ corruption. At 1\% corruption the object MAE rises by less than 0.03~mm from oracle in either frame count; at 5\% uniform corruption 1-frame PhiCalNet reaches 4.65~mm and 3-frame PhiCalNet 1.29~mm, both within 5\% of their respective clean baselines. Performance degrades materially only once corruption crosses roughly 10\%, and even at that point the 3-frame 1.54~mm MAE is still below the clean 1-frame baseline. The two frame counts exhibit essentially the same sensitivity \emph{shape} with the 3-frame curve shifted down by $\sim 4\times$, consistent with its $\sim 4\times$ better clean accuracy: more observational information raises the overall accuracy ceiling without changing how the pipeline metabolizes imperfect $k$.

The learned-$k$ variants of Section~\ref{sec:phicalnet_learnedk} achieve only 20--50\% pixel-level $k$ accuracy, placing them far beyond the right edge of this sweep, consistent with their observed 596--1439~mm MAE. Notably, their MAE exceeds even the 25\%-uniform-corruption case (5.75--8.67~mm), which is because their $k$ errors are not randomly distributed but concentrate spatially on ambiguous flat-surface regions: the learned-$k$ failure is \emph{structured} error, and the noisy-$k$ sweep underestimates the impact because uniform random corruption scatters its errors across the object rather than piling them on top of one another.

The operational implication is that PhiCalNet's sensitivity to $k$ is shallow in the regime any reasonable classical unwrap source (gray-code decoding, multi-frequency phase shifting, temporal unwrapping) will operate in. The oracle-$k$ assumption is therefore a convenient formalism rather than a strict deployment requirement: an upstream $k$-source that achieves $\geq 95\%$ per-pixel accuracy (well within reach of standard structured-light decoding pipelines) keeps PhiCalNet within 5\% of its oracle-$k$ accuracy. What the noisy-$k$ sweep does \emph{not} rescue is the ill-posedness of single-frame $k$-prediction itself, which Section~\ref{sec:phicalnet_learnedk} diagnoses as a geometric limitation rather than a fidelity issue.

\section{Mechanistic Interpretability of PhiCalNet}
\label{sec:phicalnet_interpretability}

Section~\ref{sec:interpretability} recapped the mechanistic diagnosis of the UNet baseline: shape templates are encoded more explicitly than depth values, attention concentrates on object boundaries rather than fringe patterns, and the network fails on a featureless plane that retains valid fringes. The error analysis of Section~\ref{sec:phaseunet_error_analysis} attributed PhiCalNet's residual error to wrap-boundary sign flips in the predicted wrapped phase, a localized failure mode that is mechanistically different from the diffuse shape-prior failure of the UNet baseline. The natural question is whether the same three interpretability techniques, applied to PhiCalNet, support that mechanistic distinction directly: does PhiCalNet \emph{represent} and \emph{attend to} the fringe-to-phase relationship its architecture prescribes, or has the phase-intermediate framing been bypassed by another shortcut the calibration layer happens to permit?

We therefore apply the same battery of techniques to PhiCalNet, with two adaptations needed for the phase-intermediate architecture. First, the probing target set is extended from $\{$edges, depth$\}$ to $\{$edges, depth, $\sin\phi$, $\cos\phi\}$, because phase, not depth, is the network's trained output and the asymmetry between phase decoding and depth decoding is itself diagnostic. Second, GradCAM is computed in two modes: \emph{phase mode}, in which gradients are taken with respect to the mean of the network's $(\hat{s}, \hat{c})$ output, and \emph{depth mode}, in which gradients flow through the (analytical) unwrap and calibration steps to the final depth prediction. The phase mode answers ``what does the network attend to when computing its trained output?'' and the depth mode answers ``what does it attend to when we backpropagate from the deployment output?''. The contrast between the two modes, on the same trained weights, is informative about whether the network's competence really resides at the phase stage.

\subsection{Linear Probing Analysis}
\label{sec:phicalnet_probing}

We train spatial probes (the same four-convolution architecture as in Section~\ref{sec:interpretability}, 30 epochs each) at all nine PhiCalNet layers (skip1--skip4, bottleneck, up1--up4) to predict four targets: the edge map, the ground-truth depth, $\sin\phi$, and $\cos\phi$. The phase targets are computed from the unwrapped ground-truth phase available in the calibration data and are stored as $\sin\phi, \cos\phi \in [-1, 1]$.

To compare phase probe loss with depth and edge probe loss on the same scale, we apply the bijective transform $y = (x+1)/2$ to the phase targets, mapping $(\sin\phi, \cos\phi)$ to $[0, 1]$ to match the range of edges and depth. Because the transform is linear, an unconstrained linear probe trained on the rescaled targets produces predictions that rescale by the same factor and an MSE that scales by $(1/2)^2 = 1/4$. We therefore divide the raw phase MSE by 4 to obtain the rescaled-target MSE. No re-training is required, no information is lost, and the comparison is on equal footing.

\begin{table}[pos=tbp]
\caption{PhiCalNet linear probing validation MSE loss across nine layers and four targets. Phase losses are reported on the $[0,1]$-rescaled scale (raw $[-1,1]$ MSE divided by 4) for direct comparison with edges and depth. Lower is better. The right two columns give depth/edges and depth/$\sin\phi$ ratios.}
\label{tab:phicalnet_probing_results}

\centering
\small
\resizebox{\ifdim\width>\columnwidth\columnwidth\else\width\fi}{!}{%
\begin{tabular}{@{}lcccccc@{}}
\toprule
\textbf{Layer} & \textbf{Edges} & \textbf{Depth} & \textbf{$\sin\phi$} & \textbf{$\cos\phi$} & \textbf{D/E} & \textbf{D/$\sin\phi$} \\
\midrule
enc1 (skip1) & 0.000348 & 0.002235 & 0.000306 & 0.000279 & 6.42 & 7.30 \\

enc2 (skip2) & 0.000345 & 0.001865 & 0.000219 & 0.000241 & 5.41 & 8.51 \\

enc3 (skip3) & 0.000356 & 0.001703 & 0.000165 & 0.000183 & 4.78 & 10.32 \\

enc4 (skip4) & 0.000390 & 0.001631 & 0.000276 & 0.000219 & 4.18 & 5.91 \\

bottleneck   & 0.000463 & 0.002223 & 0.000758 & 0.000732 & 4.80 & 2.93 \\

dec1 (up1)   & 0.000431 & 0.001890 & 0.000376 & 0.000386 & 4.38 & 5.03 \\

dec2 (up2)   & 0.000388 & 0.001838 & 0.000248 & 0.000245 & 4.74 & 7.41 \\

dec3 (up3)   & 0.000407 & 0.002161 & 0.000260 & 0.000245 & 5.31 & 8.31 \\

dec4 (up4)   & 0.000416 & 0.002244 & 0.000297 & 0.000272 & 5.39 & 7.56 \\

\textbf{Average} & \textbf{0.000394} & \textbf{0.001977} & \textbf{0.000323} & \textbf{0.000311} & \textbf{5.02} & \textbf{6.12} \\
\bottomrule
\end{tabular}}
\end{table}

\begin{figure}[pos=tbp]
    \centering
    \includegraphics[width=0.95\linewidth]{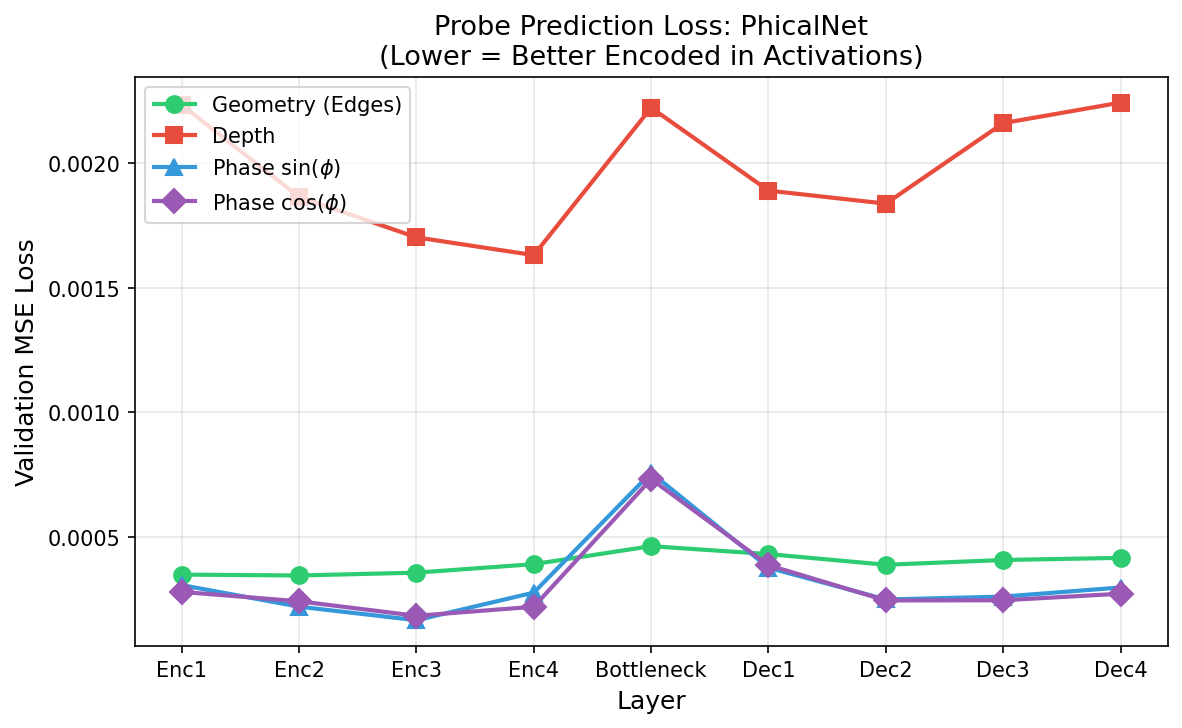}
    \caption{PhiCalNet linear probing analysis across all four targets. Phase ($\sin\phi$, $\cos\phi$) is the most decodable target everywhere except the bottleneck and is best decoded at \texttt{enc3}. Depth is the least decodable target by an order of magnitude, despite being the network's deployment output. The bottleneck spike for the phase targets is a resolution artifact (bottleneck activations live at $7.5 \times 7.5$ and the probe target is at $30 \times 30$, so high-frequency fringe content is blurred by the upsampling); the recovery at \texttt{dec2} confirms that the decoder reinjects fringe detail through the skip connections.}
    \label{fig:phicalnet_probing_analysis}
\end{figure}

The probing profile inverts the UNet baseline. Table~\ref{tab:phicalnet_probing_results} and Fig.~\ref{fig:phicalnet_probing_analysis} report the full results. For UNet (Table~\ref{tab:probing_results}, average D/E ratio 2.82), edges are the most decodable target overall, the depth/edges asymmetry tightens in the decoder where the network actually predicts depth, and depth is encoded most explicitly at \texttt{dec2}--\texttt{dec3}, indicating that depth-as-shape is consolidated in the decoder. For PhiCalNet, phase is the most decodable target in eight of nine layers, the encoder layer \texttt{enc3} achieves the lowest phase MSE in the network ($1.65 \times 10^{-4}$ for $\sin\phi$, $1.83 \times 10^{-4}$ for $\cos\phi$), and depth probe loss remains roughly $5\times$ above edge loss and $6\times$ above phase loss \emph{everywhere}. Three observations follow.

First, the network has no explicit internal representation of depth. This is consistent with the architecture by design: depth is the deterministic output of the calibration layer applied to predicted phase, and the network has no incentive to allocate features to a quantity it never produces directly. The probing result is mechanistic confirmation that the calibration layer carries the geometric transform rather than learned weights. Second, phase encoding is consolidated in the encoder (best at \texttt{enc3}), not in the decoder, which is the opposite of the UNet baseline's depth-as-shape consolidation. The decoder's role in PhiCalNet is to upsample and refine an already-formed phase representation, not to construct a target representation from scratch. Third, the bottleneck spike is a resolution artifact: bottleneck activations live at $7.5 \times 7.5$, the probe target lives at $30 \times 30$, and bilinear upsampling smears the high-frequency fringe content that phase encoding requires. The recovery at \texttt{dec2} (back to encoder-level phase loss) is the signature of the skip connections reinjecting fringe detail at the appropriate spatial resolution.

\subsection{Dual-Mode GradCAM Analysis}
\label{sec:phicalnet_gradcam}

We compute GradCAM heatmaps at six layers (\texttt{enc3}, \texttt{enc4}, bottleneck, \texttt{dec1}, \texttt{dec3}, \texttt{dec4}) over the same 30-test-sample subset used for the UNet GradCAM analysis, in two modes:
\begin{itemize}
    \item \textbf{Phase mode}: the gradient target is $\frac{1}{2}\bigl(\overline{\hat{s}} + \overline{\hat{c}}\bigr)$, the mean of the predicted $(\sin\phi, \cos\phi)$ output.
    \item \textbf{Depth mode}: the gradient target is $\overline{\hat{d}}$, the mean of the calibrated depth output, computed by passing the predicted $(\hat{s}, \hat{c})$ through the analytical unwrap (with oracle $k$) and the calibration layer.
\end{itemize}
The heatmaps are then correlated, pixel-wise, with two reference maps: a Sobel edge map of the ground-truth depth and a local-standard-deviation fringe-intensity map. Phase mode answers ``what does PhiCalNet attend to when computing its trained quantity?'' and depth mode answers ``what does it attend to when we backpropagate from the derived deployment quantity?''. The contrast between modes is what isolates the role of the calibration layer.

\begin{table}[pos=tbp]
\caption{PhiCalNet GradCAM correlation analysis, dual-mode. Each mode is computed on the same trained weights and the same 30-sample test subset. Phase-mode gradient target is $\overline{\hat{s}} + \overline{\hat{c}}$; depth-mode gradient target is the calibrated depth output. The ratio column is r(CAM, Edges) / r(CAM, Fringes); ratio close to 1 means equal attention to edges and fringes, ratio $>1$ means edge-favored.}
\label{tab:phicalnet_gradcam_results}

\centering
\small
\resizebox{\ifdim\width>\columnwidth\columnwidth\else\width\fi}{!}{%
\begin{tabular}{@{}lcccccc@{}}
\toprule
& \multicolumn{3}{c}{\textbf{Phase mode}} & \multicolumn{3}{c}{\textbf{Depth mode}} \\
\cmidrule(lr){2-4}\cmidrule(lr){5-7}
\textbf{Layer} & \textbf{r(Edge)} & \textbf{r(Fringe)} & \textbf{Ratio} & \textbf{r(Edge)} & \textbf{r(Fringe)} & \textbf{Ratio} \\
\midrule
enc3       &     0.132 &     0.032 &     4.16 &     0.092 &     0.007 &     9.16 \\

enc4       &     0.155 &     0.164 &     0.94 &     0.148 &     0.128 &     1.16 \\

bottleneck &     0.107 &     0.067 &     1.59 &     0.070 &     0.044 &     1.60 \\

dec1       &     0.055 &     0.016 &     3.53 &     0.071 &     0.038 &     1.84 \\

dec3       & $-$0.053  & $-$0.035  & $-$5.30  &     0.111 &     0.061 &     1.82 \\

dec4       &     0.142 &     0.262 &     0.54 &     0.135 &     0.129 &     1.04 \\

\textbf{Average} & \textbf{0.090} & \textbf{0.084} & \textbf{1.06} & \textbf{0.104} & \textbf{0.068} & \textbf{1.54} \\
\bottomrule
\end{tabular}}
\end{table}

\begin{figure*}[pos=tp]
    \centering
    \begin{subfigure}[t]{\linewidth}
        \centering
        \includegraphics[width=\linewidth]{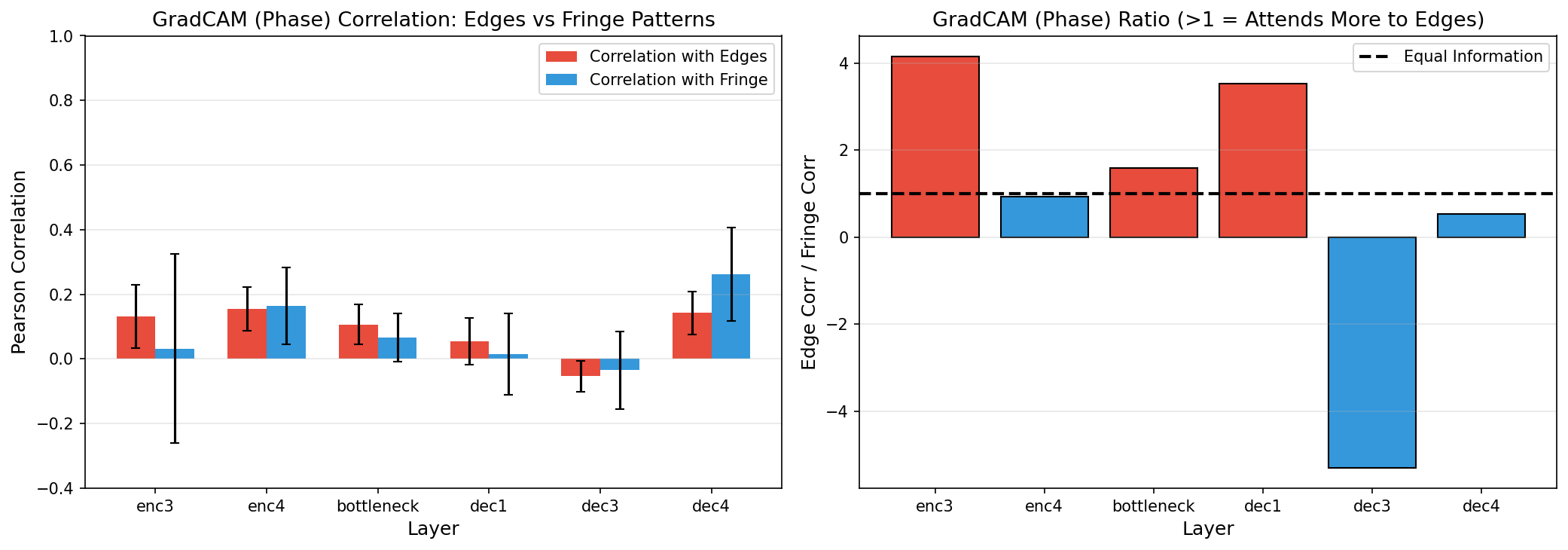}
        \caption{Phase mode}
    \end{subfigure}
    \\[0.6em]
    \begin{subfigure}[t]{\linewidth}
        \centering
        \includegraphics[width=\linewidth]{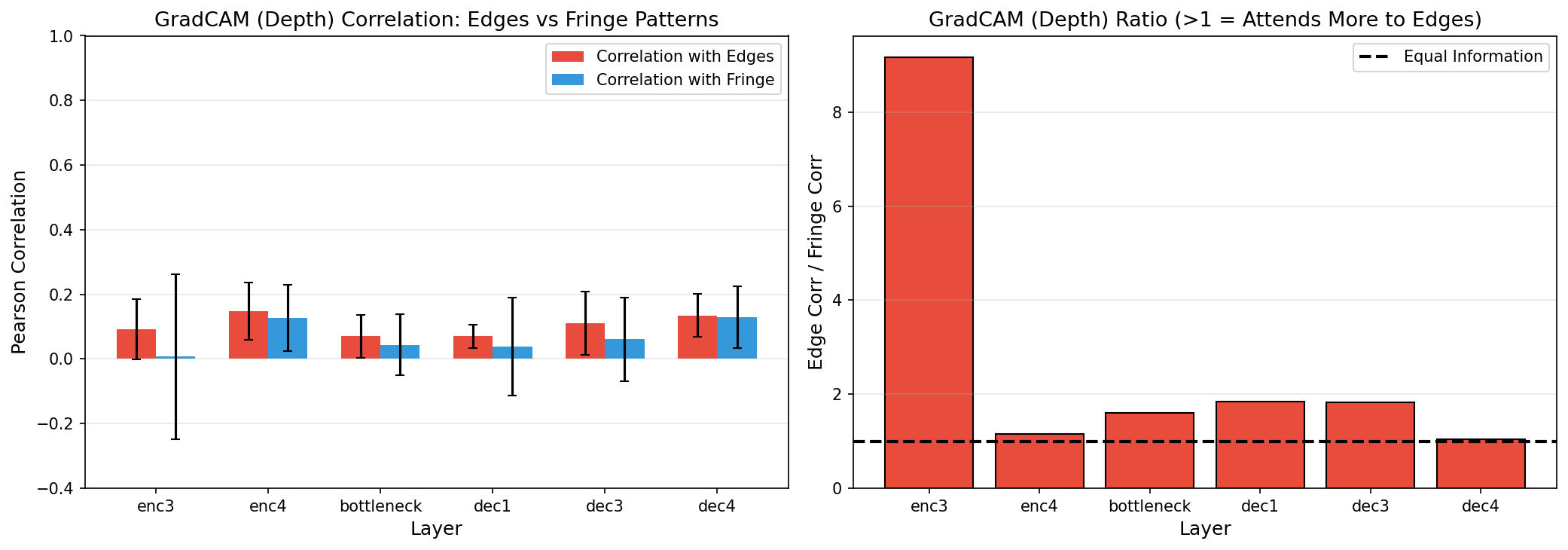}
        \caption{Depth mode}
    \end{subfigure}
    \caption{PhiCalNet GradCAM correlation summary in the two modes. Phase mode (top) gives an average edge/fringe ratio of 1.06, with fringe correlation matching or exceeding edge correlation at the layers that drive the output (\texttt{enc4}, \texttt{dec4}). Depth mode (bottom) gives an average ratio of 1.54, edge-favored, because gradients flowing back through the analytical calibration layer inherit the region/extent sensitivity of the depth output. The shift between modes, on the same trained weights, isolates the role of the calibration layer in introducing edge bias to the depth output without that bias being present in the network's phase representation.}
    \label{fig:phicalnet_gradcam_summary}
\end{figure*}

\begin{figure*}[pos=tp]
    \centering
    (a) Phase mode\par\medskip
    \begin{subfigure}[t]{0.13\linewidth}
        \centering
        \includegraphics[width=\linewidth]{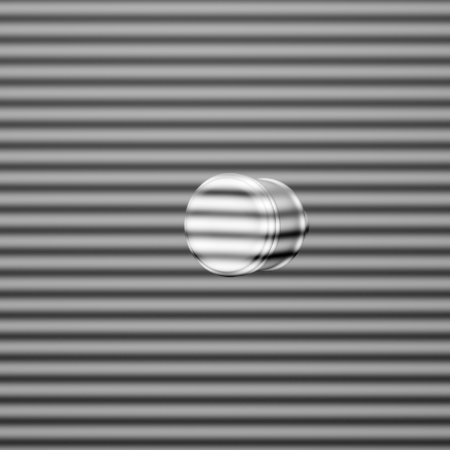}
        \caption{Input}
    \end{subfigure}\hfill
    \begin{subfigure}[t]{0.13\linewidth}
        \centering
        \includegraphics[width=\linewidth]{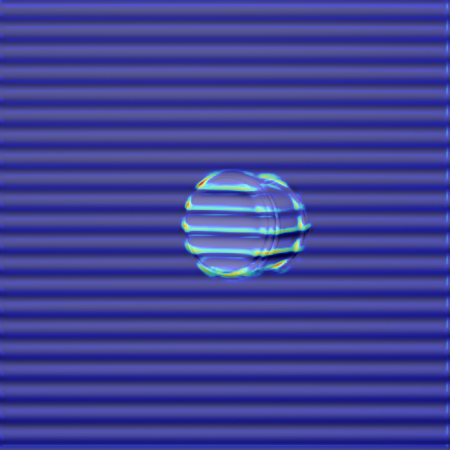}
        \caption{enc3}
    \end{subfigure}\hfill
    \begin{subfigure}[t]{0.13\linewidth}
        \centering
        \includegraphics[width=\linewidth]{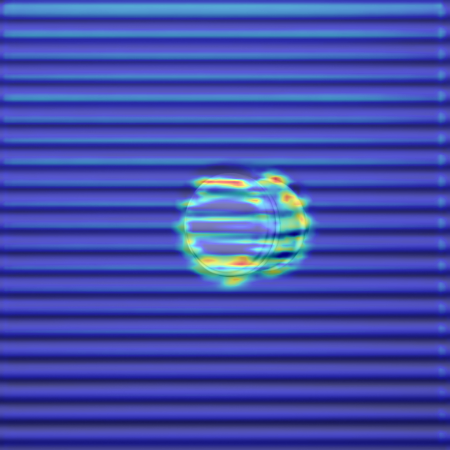}
        \caption{enc4}
    \end{subfigure}\hfill
    \begin{subfigure}[t]{0.13\linewidth}
        \centering
        \includegraphics[width=\linewidth]{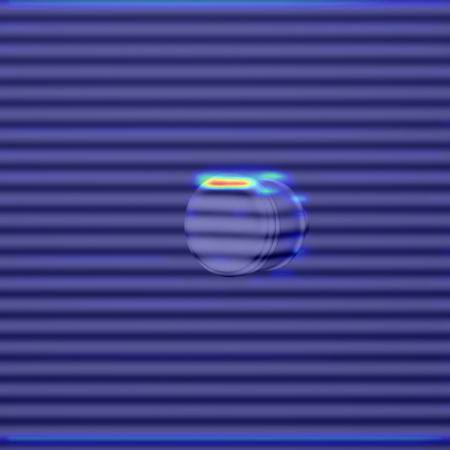}
        \caption{bottleneck}
    \end{subfigure}\hfill
    \begin{subfigure}[t]{0.13\linewidth}
        \centering
        \includegraphics[width=\linewidth]{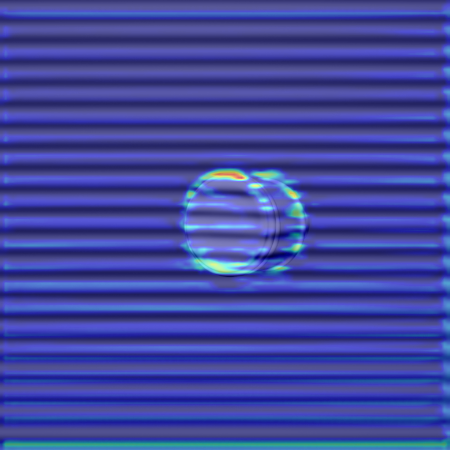}
        \caption{dec1}
    \end{subfigure}\hfill
    \begin{subfigure}[t]{0.13\linewidth}
        \centering
        \includegraphics[width=\linewidth]{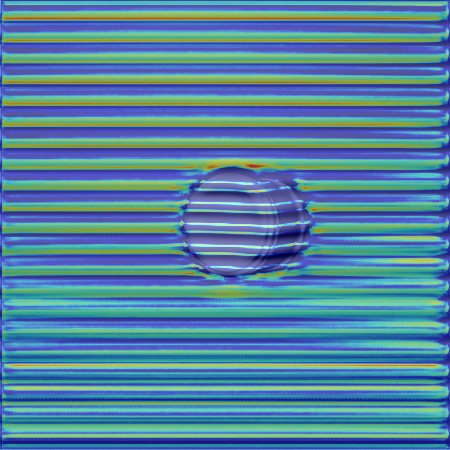}
        \caption{dec3}
    \end{subfigure}\hfill
    \begin{subfigure}[t]{0.13\linewidth}
        \centering
        \includegraphics[width=\linewidth]{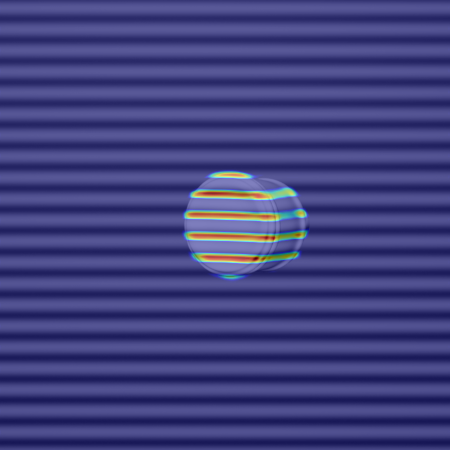}
        \caption{dec4}
    \end{subfigure}
    \\[0.6em]
    (b) Depth mode\par\medskip
    \begin{subfigure}[t]{0.13\linewidth}
        \centering
        \includegraphics[width=\linewidth]{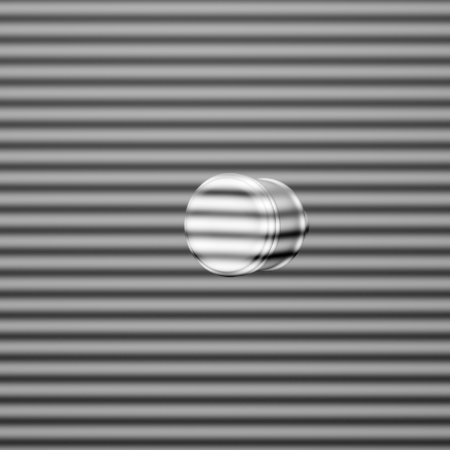}
        \caption{Input}
    \end{subfigure}\hfill
    \begin{subfigure}[t]{0.13\linewidth}
        \centering
        \includegraphics[width=\linewidth]{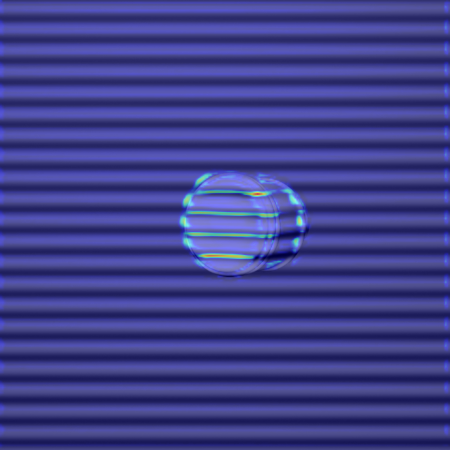}
        \caption{enc3}
    \end{subfigure}\hfill
    \begin{subfigure}[t]{0.13\linewidth}
        \centering
        \includegraphics[width=\linewidth]{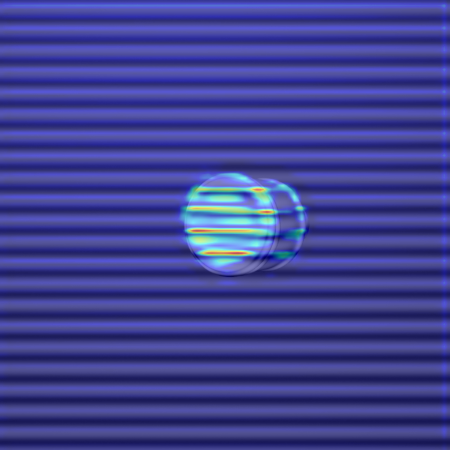}
        \caption{enc4}
    \end{subfigure}\hfill
    \begin{subfigure}[t]{0.13\linewidth}
        \centering
        \includegraphics[width=\linewidth]{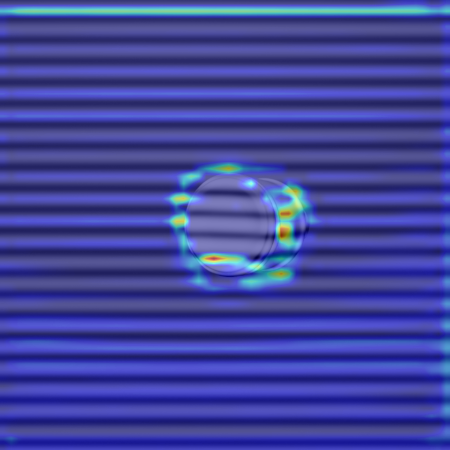}
        \caption{bottleneck}
    \end{subfigure}\hfill
    \begin{subfigure}[t]{0.13\linewidth}
        \centering
        \includegraphics[width=\linewidth]{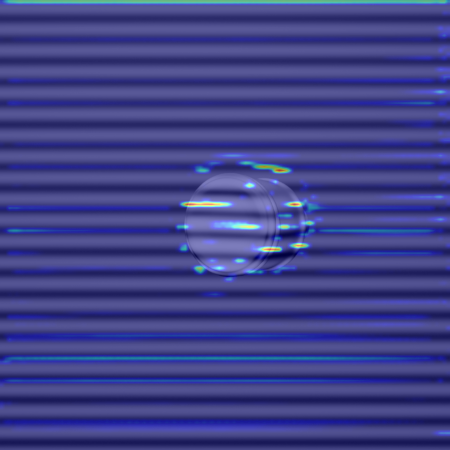}
        \caption{dec1}
    \end{subfigure}\hfill
    \begin{subfigure}[t]{0.13\linewidth}
        \centering
        \includegraphics[width=\linewidth]{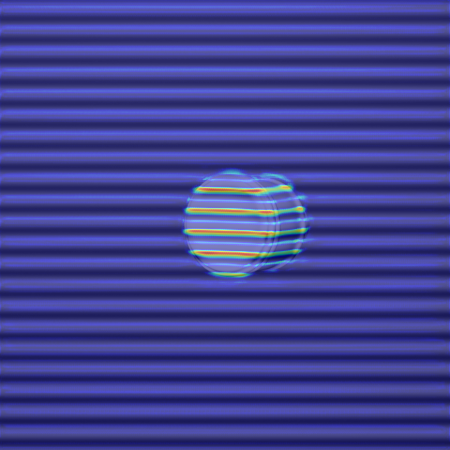}
        \caption{dec3}
    \end{subfigure}\hfill
    \begin{subfigure}[t]{0.13\linewidth}
        \centering
        \includegraphics[width=\linewidth]{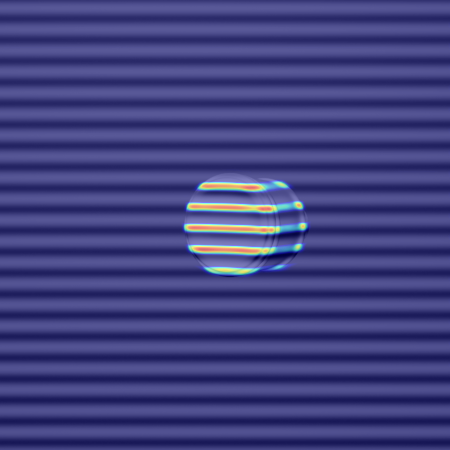}
        \caption{dec4}
    \end{subfigure}
    \caption{PhiCalNet GradCAM visualization for test sample \texttt{container\_bottle\_A60} in the two modes. Phase mode (top row, gradient target $= \overline{\hat{s}} + \overline{\hat{c}}$) and depth mode (bottom row, gradient target $= $ calibrated depth) are produced from identical activations and weights; the difference between rows is attributable to the gradient target alone. Phase-mode attention covers the fringes within and around the object body; depth-mode attention concentrates more on the object body and its boundary, and the bottleneck panel in the depth row in particular shows a region/extent-style activation that the phase row does not exhibit, consistent with the calibration layer's smooth dependence on the global phase field introducing extent sensitivity to the depth output.}
    \label{fig:phicalnet_gradcam_sample}
\end{figure*}

The phase-mode and depth-mode columns of Table~\ref{tab:phicalnet_gradcam_results} should be read together with attention to per-layer noise levels (Fig.~\ref{fig:phicalnet_gradcam_sample} shows a representative sample where the mode contrast is visible directly on the activations). Layers with $|r| < 0.1$ for both reference maps (\texttt{dec3} in phase mode, with r values $-0.053$ and $-0.035$) sit inside the noise band of pixel-correlations on $30 \times 30$ heatmaps; the apparent ratio of $-5.30$ at \texttt{dec3} is small-over-small noise rather than a meaningful anti-correlation, and the corresponding bar is plotted in Fig.~\ref{fig:phicalnet_gradcam_summary} for completeness but should be interpreted as ``not aligned with either reference map'' rather than as evidence of attention to background regions. The interpretively load-bearing layers are those with both correlations well above the $|r| \approx 0.1$ noise floor: \texttt{enc4} (phase ratio 0.94, depth ratio 1.16), bottleneck (1.59 / 1.60), and \texttt{dec4} (0.54 / 1.04).

Two consistent patterns emerge. First, in phase mode, fringe correlation matches or exceeds edge correlation at the layers that actually drive the output: \texttt{enc4} is essentially balanced (ratio 0.94), and \texttt{dec4}, the deepest decoder layer immediately before the phase head, is fringe-favored at ratio 0.54 with the largest absolute fringe correlation in the network ($r = 0.262$). At the boundary between the network's last-stage spatial computation and the analytical calibration layer, the network is attending more to fringe pattern than to object edges. This is the cleanest fringe-attention signature recovered from a network in this study and the qualitative inverse of the UNet baseline's \texttt{dec4} ratio of 1.38 (Table~\ref{tab:gradcam_results}). Second, the average phase-mode ratio of 1.06 is itself meaningful when read against the UNet baseline's average ratio of 1.28 on the same test split: the same six-layer GradCAM pipeline, applied to a phase-trained network, produces a saliency profile in which fringes and edges are roughly co-equal contributors to the phase output, with the edge contribution becoming dominant only when the gradient target is moved through the calibration layer.

The depth-mode column (average ratio 1.54) shifts the same activations toward edge-favored saliency. The mechanism is straightforward: the calibration layer is a smooth function of the network's full predicted phase field, and the calibrated depth output depends on the global integral of phase over the image extent, with more sensitivity to where the object \emph{is} (its spatial extent) than to local fringe variation within it. Gradients backpropagated through this transform therefore inherit a region/extent sensitivity that the underlying activations do not have when read directly. The shift from a phase-mode average of 1.06 to a depth-mode average of 1.54, on the same trained weights, is the empirical signature of this effect; if the network were silently using shape priors to produce phase, the two modes would not separate. The contrast is the result.

\subsection{Flat Plane Out-of-Distribution Test}
\label{sec:phicalnet_flat}

\begin{figure*}[pos=tp]
    \centering
    \begin{subfigure}[c]{0.20\linewidth}
        \centering
        \includegraphics[width=\linewidth]{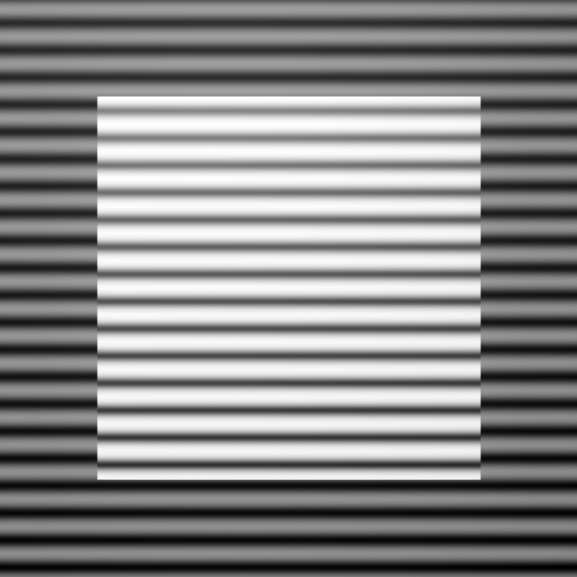}
        \caption{Input fringe image}
    \end{subfigure}\hfill
    \begin{subfigure}[c]{0.255\linewidth}
        \centering
        \includegraphics[width=\linewidth]{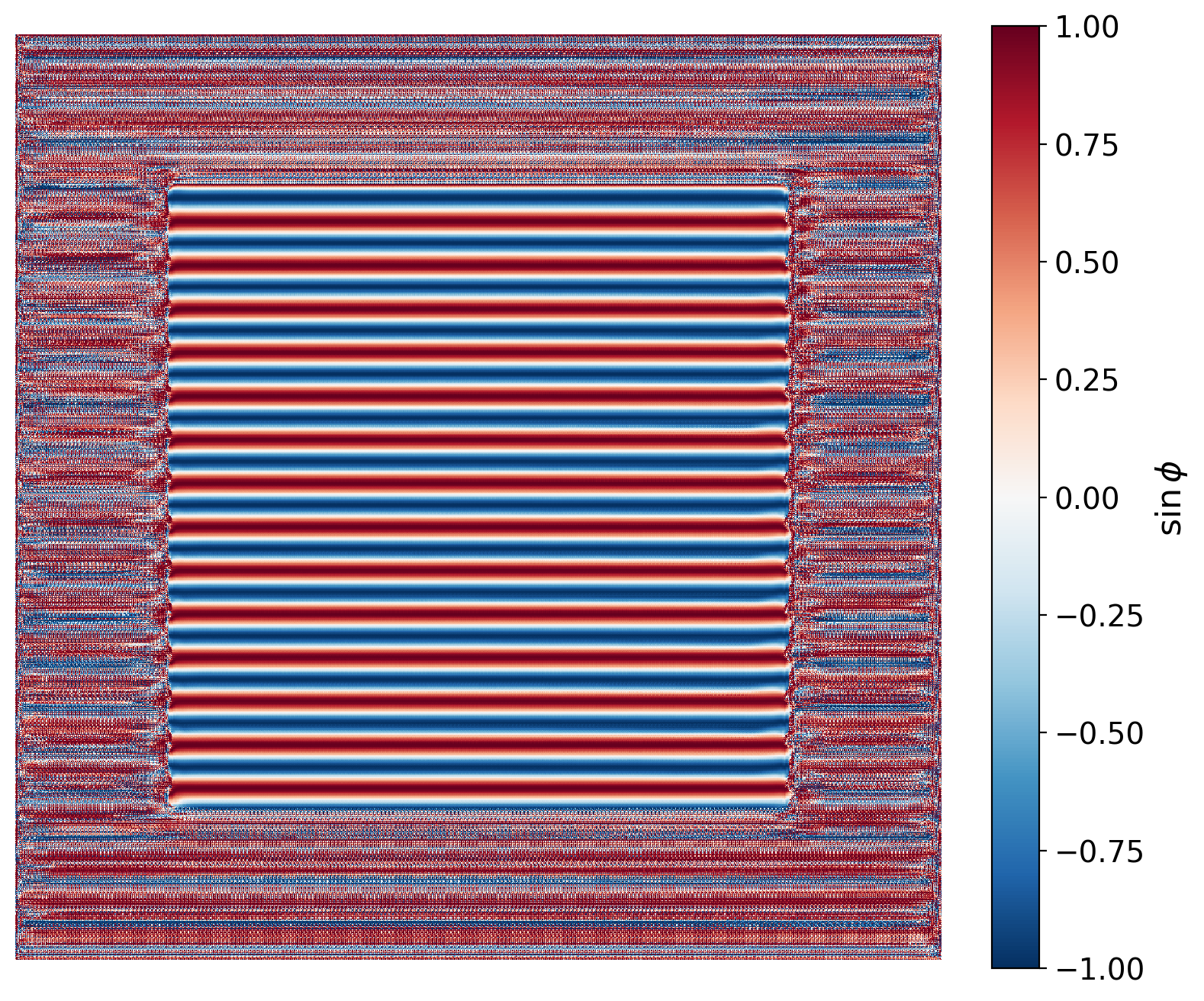}
        \caption{Predicted $\sin\phi$}
    \end{subfigure}\hfill
    \begin{subfigure}[c]{0.255\linewidth}
        \centering
        \includegraphics[width=\linewidth]{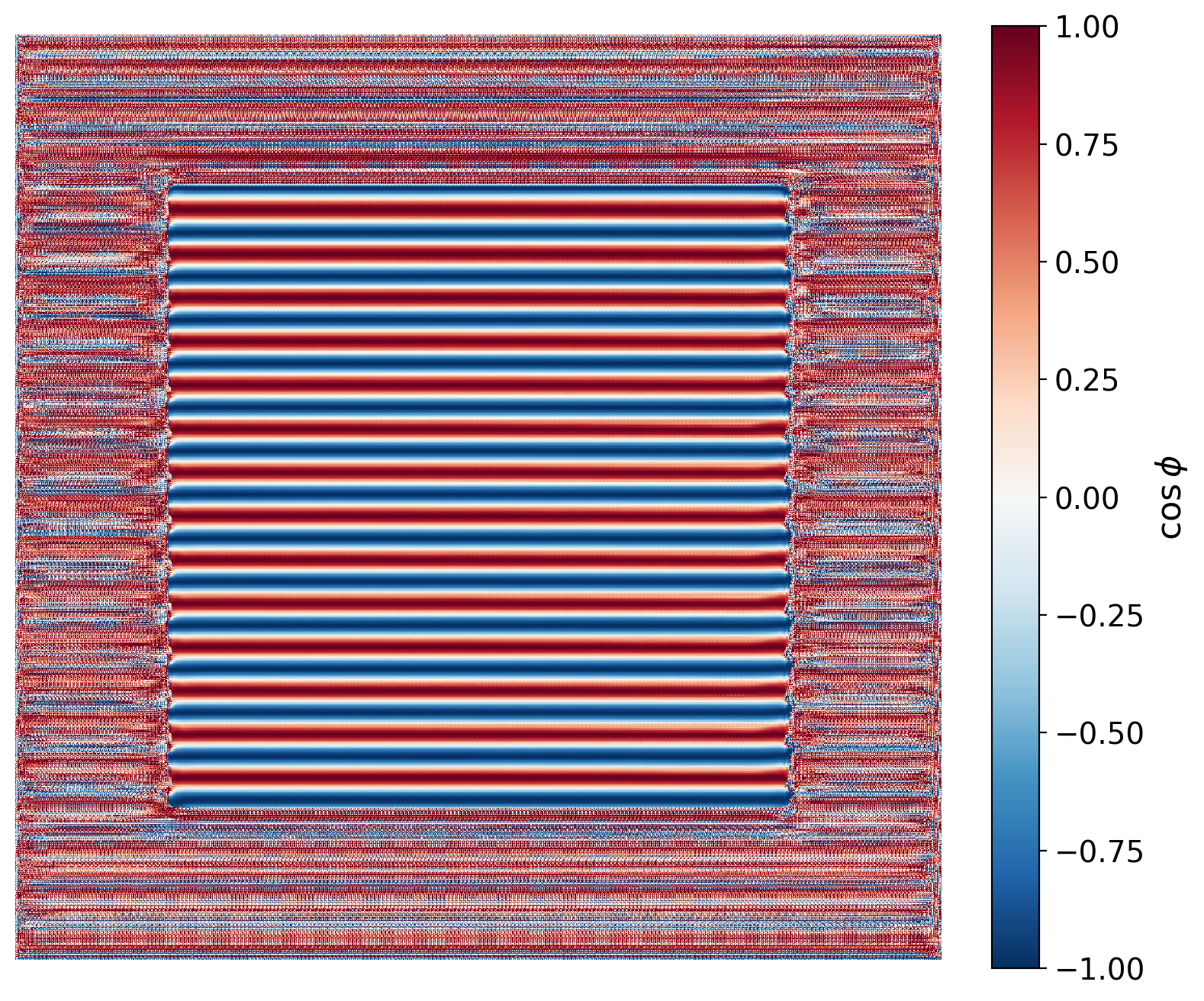}
        \caption{Predicted $\cos\phi$}
    \end{subfigure}\hfill
    \begin{subfigure}[c]{0.255\linewidth}
        \centering
        \includegraphics[width=\linewidth]{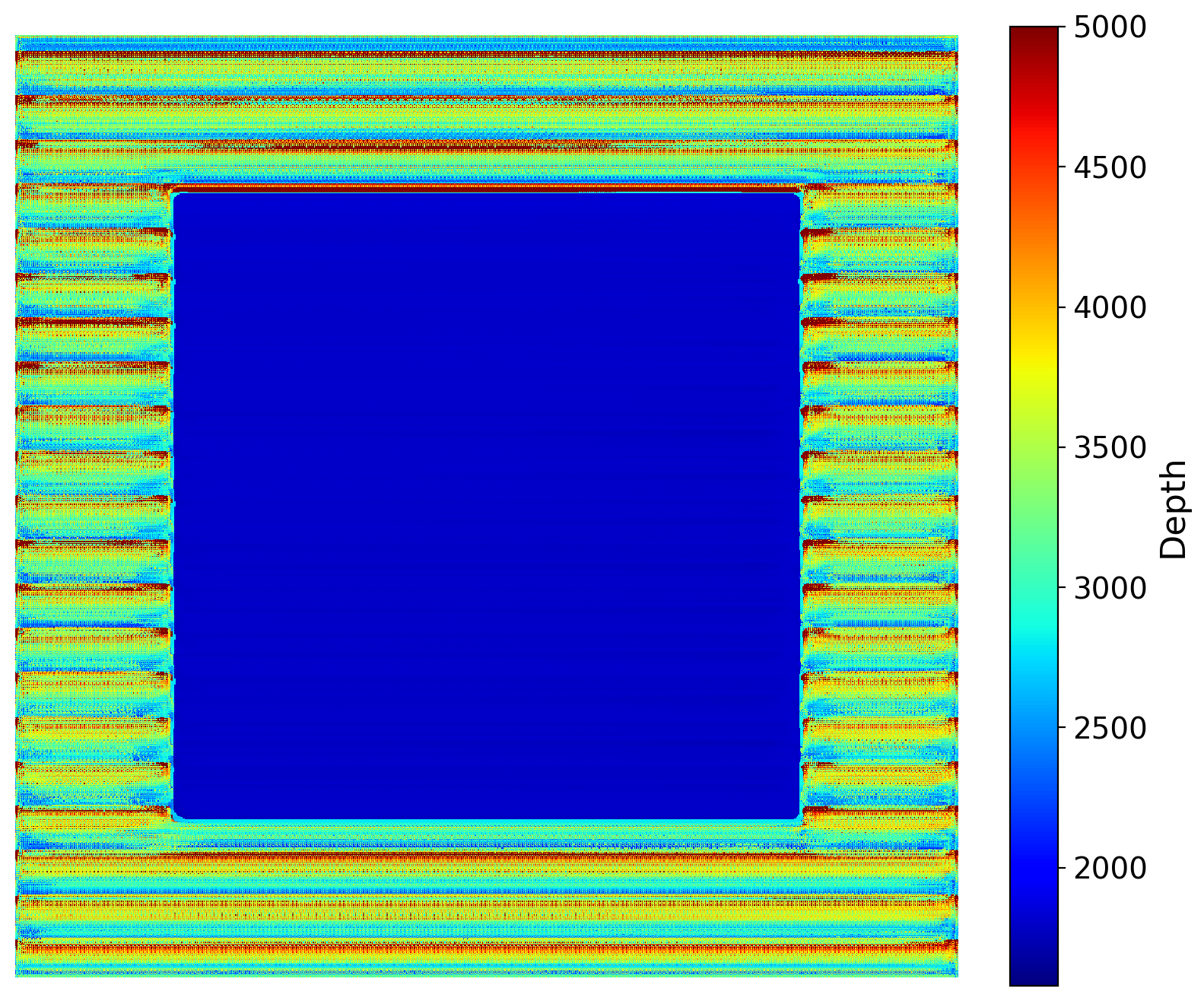}
        \caption{Depth prediction (range: [1579, 5000]~mm)}
    \end{subfigure}
    \caption{PhiCalNet flat-plane out-of-distribution test. (a) Input fringe image of a flat plane captured at 1.8~m, within the trained depth range. (b, c) Predicted $\sin\phi$ and $\cos\phi$ recover the fringe pattern everywhere fringes are visible, including over the background the network never saw at train time. The quarter-period offset between their bands is the expected sin/cos quadrature, confirming the outputs do not collapse onto each other, while the vertical-bar artifacts at the foreground boundary reflect a residual in-distribution prior. (d) Depth, from unwrapping the predicted phase with the oracle fringe order and the calibration layer, is essentially uniform across the plane at 1795~mm (standard deviation 6~mm), correctly placing the surface in the trained range; the larger off-plane values arise only over the unseen background, not the plane.}
    \label{fig:phicalnet_flat_plane_phase}
\end{figure*}

We apply the same flat-plane harness used for the UNet baseline (Section~\ref{sec:interpretability}, plane captured at 1.8~m within the trained depth range). PhiCalNet's primary output is the wrapped phase, so we read the predicted $(\hat{s}, \hat{c})$ field directly as the main diagnostic and additionally carry it through the fixed unwrap and calibration layers, using the oracle fringe order as in the rest of the PhiCalNet pipeline, to obtain the calibrated depth. Figure~\ref{fig:phicalnet_flat_plane_phase} shows the input fringe, both predicted phase components, and the resulting depth. Reading the phase directly isolates what the network actually produces, while the depth shows whether that phase, once calibrated, places the flat surface at the correct distance, which is the question the flat-plane test was designed to answer.

The phase prediction is the load-bearing observation. Across the full image, including the gray background fringes that lie outside the training distribution, both $\hat{s}$ and $\hat{c}$ recover a coherent horizontal-fringe field, and they do so in the correct quadrature: the bright bands of $\hat{c}$ sit at the zero crossings of $\hat{s}$ and vice versa, which is the analytic relationship $\sin\phi$ and $\cos\phi$ should satisfy and is a stronger consistency check than reading either component alone, since a network producing one component as a learned shortcut from the input intensity could not maintain the quadrature offset against the other. Vertical-bar artifacts at the boundary of the rectangular foreground are present, confirming that the rectangular-foreground prior of the training distribution has not fully washed out, but the phase signal is recovered everywhere fringes are visible. By contrast, the UNet baseline on the same flat-plane input collapsed its sole output (depth) to a near-zero background value across the entire surface (Section~\ref{sec:interpretability}; corresponding figure in \citealp{haroon2026diagnosis}). The diagnostic question of the flat-plane test was whether the network has the capacity to extract depth information from a fringe pattern that lacks its training-distribution shape cues; PhiCalNet retains that capacity at the phase stage, the UNet baseline does not.

Carrying the phase through the calibration layer makes the contrast quantitative. With the oracle fringe order supplied as elsewhere in PhiCalNet, the predicted depth is essentially uniform across the plane at 1795~mm with a standard deviation of 6~mm (Fig.~\ref{fig:phicalnet_flat_plane_phase}d), within a few millimeters of the true 1.8~m distance and comfortably inside the 1.5--2.1~m trained range. The network therefore recognizes that a flat plane should produce flat depth, which is exactly the diagnostic the test was designed to probe and exactly what the UNet baseline failed. The much larger values that stretch the full-field range toward 5000~mm fall entirely on the unseen background beyond the rectangular foreground, where the residual foreground prior and wrap-order ambiguity corrupt the unwrap; they do not belong to the plane surface itself. The depth output therefore reaches the same verdict as the phase field, at the quantity the application ultimately cares about: PhiCalNet recovers the correct depth for the flat surface, whereas the UNet baseline collapsed the identical plane to a near-zero background value across its entire surface.

\begin{figure*}[pos=tp]
    \centering
    \begin{subfigure}[t]{0.13\linewidth}
        \centering
        \includegraphics[width=\linewidth]{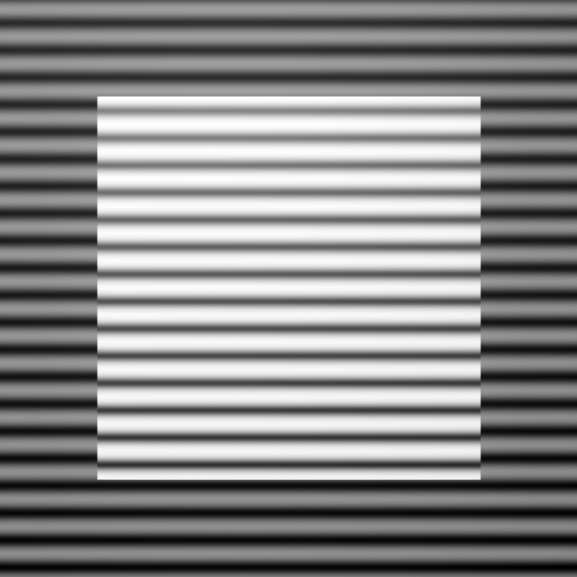}
        \caption{Input}
    \end{subfigure}\hfill
    \begin{subfigure}[t]{0.13\linewidth}
        \centering
        \includegraphics[width=\linewidth]{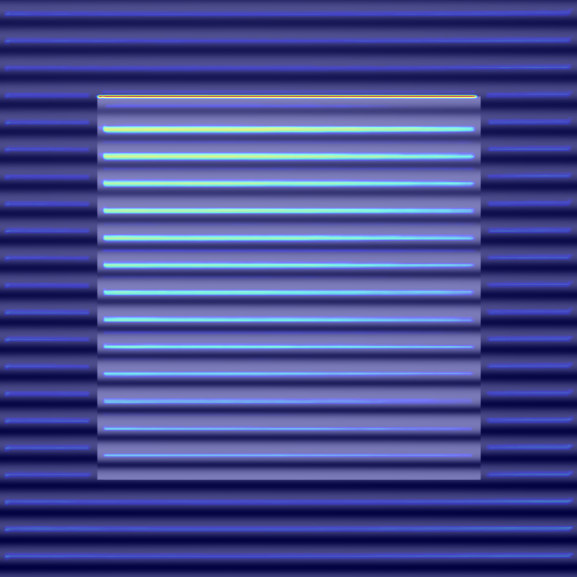}
        \caption{enc3}
    \end{subfigure}\hfill
    \begin{subfigure}[t]{0.13\linewidth}
        \centering
        \includegraphics[width=\linewidth]{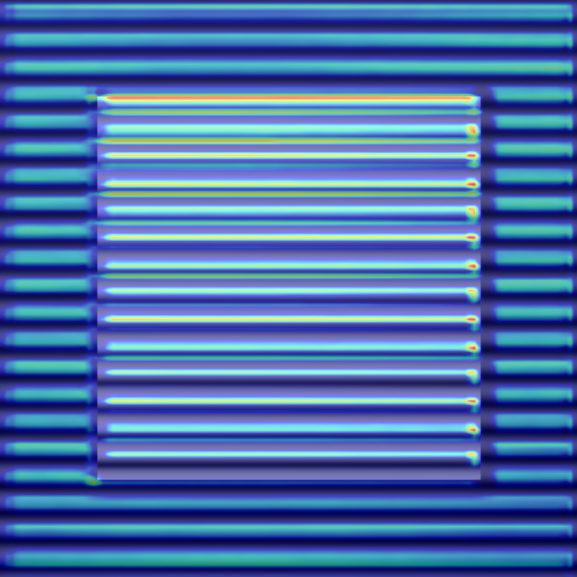}
        \caption{enc4}
    \end{subfigure}\hfill
    \begin{subfigure}[t]{0.13\linewidth}
        \centering
        \includegraphics[width=\linewidth]{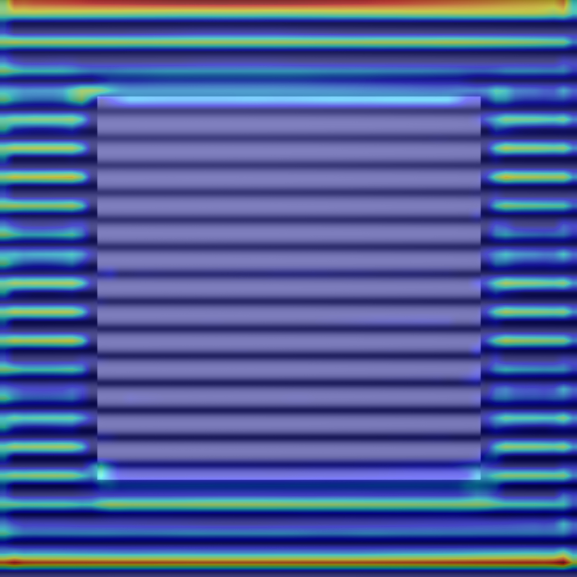}
        \caption{bottleneck}
    \end{subfigure}\hfill
    \begin{subfigure}[t]{0.13\linewidth}
        \centering
        \includegraphics[width=\linewidth]{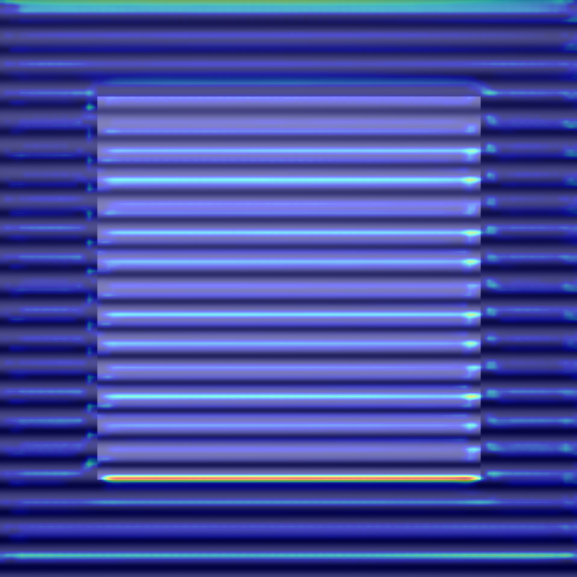}
        \caption{dec1}
    \end{subfigure}\hfill
    \begin{subfigure}[t]{0.13\linewidth}
        \centering
        \includegraphics[width=\linewidth]{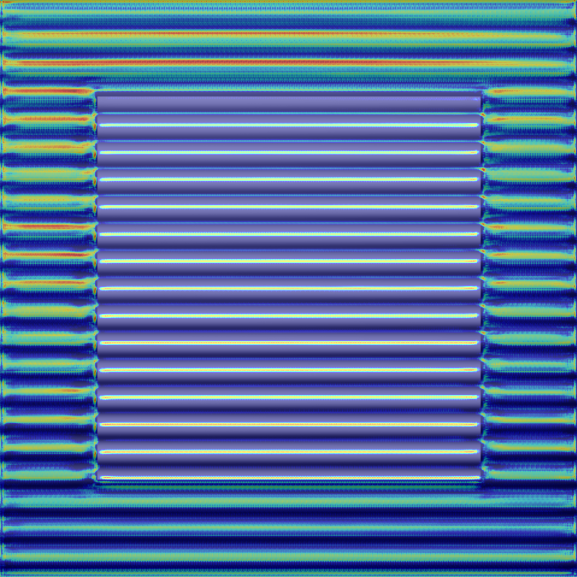}
        \caption{dec3}
    \end{subfigure}\hfill
    \begin{subfigure}[t]{0.13\linewidth}
        \centering
        \includegraphics[width=\linewidth]{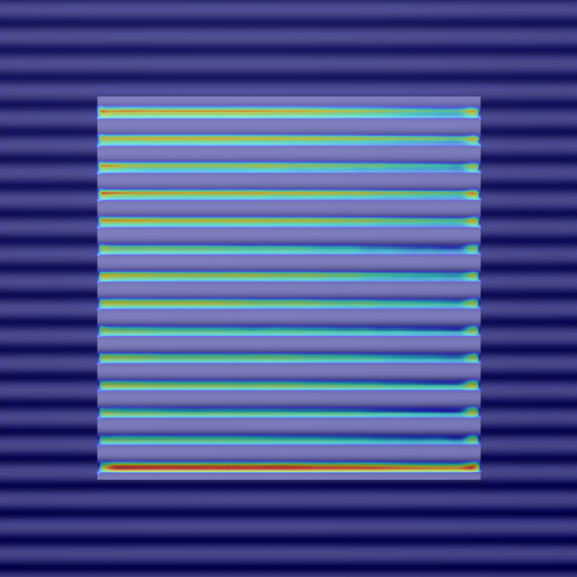}
        \caption{dec4}
    \end{subfigure}
    \caption{PhiCalNet GradCAM analysis on the flat-plane input, phase mode. Encoder layers (\texttt{enc3}, \texttt{enc4}) and decoder layers (\texttt{dec1}, \texttt{dec3}, \texttt{dec4}) both overlay attention on the horizontal fringe pattern across the full image, including the gray background fringes that lie outside the training distribution. Compare with the UNet baseline (Section~\ref{sec:interpretability}; figure in \citealp{haroon2026diagnosis}), where encoder attention selectively concentrated on the fringes inside the rectangular foreground and decoder layers showed minimal spatial attention.}
    \label{fig:phicalnet_flat_plane_gradcam}
\end{figure*}

Fig.~\ref{fig:phicalnet_flat_plane_gradcam} shows GradCAM in phase mode at all six layers on the flat-plane input: encoder attention engages with fringes both within and outside the rectangular foreground rather than collapsing onto the boundary, and decoder layers continue to overlay attention on the fringe pattern across the field. This is a different qualitative behavior from the UNet baseline's flat-plane GradCAM (Section~\ref{sec:interpretability}; figure in \citealp{haroon2026diagnosis}), where encoder attention concentrated on the fringes \emph{inside} the rectangular foreground while the decoder produced no usable spatial prediction at all.

\subsection{Implications: Physics-Based vs Geometric-Shortcut Decoding}
\label{sec:phicalnet_mi_implications}

The three analyses converge on a mechanistic explanation that is the structural inverse of the UNet baseline's:

\begin{itemize}
    \item \textbf{Linear probing} establishes that PhiCalNet's most decodable internal feature is phase, not depth, and that phase encoding is consolidated in the encoder (best at \texttt{enc3}). Depth probe loss is uniformly $5$--$6\times$ higher than phase or edge loss, indicating that the network does not maintain an explicit internal representation of depth: depth is the deterministic output of the calibration layer applied to predicted phase, and the network's learned content is at the phase stage. This is the inverse of the UNet baseline's profile, in which depth-as-shape was consolidated in the decoder (Section~\ref{sec:interpretability}, Table~\ref{tab:probing_results}).
    \item \textbf{Dual-mode GradCAM} shows that, on the same trained weights, phase-mode attention (gradient target = predicted $(\sin\phi, \cos\phi)$) gives an edge/fringe ratio of 1.06 with fringes equal-or-favored at the layers that drive the phase output, while depth-mode attention (gradient target = calibrated depth) gives a ratio of 1.54 with edges favored. The shift between modes is the empirical signature of the calibration layer introducing region/extent sensitivity to the depth output without that sensitivity being present in the underlying activations. The UNet baseline, in which the only available gradient target is depth, has a ratio of 1.28 with edges favored throughout the network.
    \item \textbf{Flat plane testing} shows that PhiCalNet recovers the fringe pattern at the phase stage on a featureless plane lacking its training-distribution shape cues, with $\hat{s}$ and $\hat{c}$ both producing coherent fringe fields that maintain their analytic quadrature offset across the full image, including over the gray background that lies outside the training distribution; carried through the calibration layer, the same phase field places the surface at a uniform 1795~mm (standard deviation 6~mm) within a few millimeters of the true 1.8~m distance, so the network not only encodes phase correctly but also recovers the correct metric depth on a stimulus that contains no shape cues. The UNet baseline collapsed its sole output to near-zero background across the same plane and could not have done otherwise: the network's only learned mapping is from shape templates to depth, and a featureless plane has no shape template to map onto.
\end{itemize}

These three pieces of evidence are mutually reinforcing in the same way as the UNet evidence was for the geometric-shortcut diagnosis, but they support the opposite conclusion. PhiCalNet's measurable internal content is the phase-from-fringe relationship its architecture prescribes, and the calibration layer carries the deterministic phase-to-depth transform without learned weights. This explains the $3.3\times$ object-MAE gap relative to the UNet baseline ($4.46$~mm vs.\ $14.54$~mm in object MAE) on physical rather than capacity-based grounds: the calibration layer pins the network to the phase-to-depth physics of FPP, removing the shape-prior solution from the hypothesis space, and the residual error is the wrap-boundary artifact identified analytically in Section~\ref{sec:phaseunet_error_analysis}, which is information-theoretic in origin (a discontinuous wrap target cannot be exactly tracked by a continuous regressor) rather than a shortcut artifact. The interpretability evidence and the residual-error analysis therefore make the same statement from two directions: PhiCalNet's competence is at the phase stage, its failures are at the phase-stage discontinuity, and its depth output inherits both.

\section{Pixel-Wise Uncertainty Quantification}
\label{sec:phicalnet_uq}

The mechanistic interpretability analyses of Sections~\ref{sec:interpretability} and~\ref{sec:phicalnet_interpretability}, complemented by the residual error analysis of Section~\ref{sec:phaseunet_error_analysis}, examined the internal representations and residuals of the UNet baseline and PhiCalNet respectively, with the goal of characterizing \emph{where} and \emph{why} each model fails. Uncertainty quantification (UQ) is the natural counterpart to that analysis from the opposite direction: rather than inspecting a model's internals, UQ asks the model itself to report \emph{what it does not know} about its own predictions, on a per-pixel basis at deployment time. The two diagnostics target the same underlying object, the structure of the model's residual error, but reach it through different evidence: MI through learned features and gradients, UQ through the variability of the predictive distribution. Together they provide complementary checks on whether the failure modes identified analytically are also failure modes the model can flag empirically. We therefore apply the same UQ method to both architectures examined mechanistically above.

We use a specific UQ framework for this purpose: \emph{pixel-wise conformal prediction}~\citep{vovk2005algorithmic,angelopoulos2023conformal} on top of a snapshot ensemble. To our knowledge, this is the first application of conformal prediction to fringe projection profilometry, and one of the first to pixel-wise depth regression more broadly. The methodological argument beyond the FPP-specific result is that MI inspects internals and UQ inspects self-reported confidence, and the two should converge on the same failure population when the underlying failure mode has a clear physical locus. The conceptual idea of pairing explainability and uncertainty has small but real precedent in the broader ML literature~\citep{teplica2025sciurus,salvi2025explainability}, but FPP makes the convergence test substantially more diagnostic than in those settings: the network's failure modes have specific geometric loci (wrap boundaries are a single physical locus) rather than the diffuse linguistic or medical patterns where prior joint MI and UQ work has been done, so MI inspection of internals and UQ inspection of self-reported confidence can be compared at a single point and shown to agree. We treat this convergence as a methodological contribution beyond the FPP-specific result; the remainder of this section sets up the conformal procedure (Section~\ref{sec:conformal_intro}), the method (Section~\ref{sec:phicalnet_uq_method}), the results, and the interpretation that ties the MI and UQ diagnoses to a single locus.

UQ in deep learning conventionally distinguishes \emph{aleatoric} uncertainty (irreducible noise inherent to the measurement, e.g.\ sensor noise or stochastic projection patterns) from \emph{epistemic} uncertainty (reducible-with-more-data uncertainty arising from the model's incomplete knowledge of the true input-to-output mapping)~\citep{kendall2017uncertainties,abdar2021review}. The two are estimated differently: aleatoric uncertainty by training the model to predict its own output variance (a heteroscedastic likelihood head), epistemic uncertainty by examining the disagreement of an ensemble of models or model checkpoints~\citep{lakshminarayanan2017deepensembles}. For single-shot FPP, both sources are present: camera noise, projector blur, and the way the projected fringe pattern interacts with each material's surface reflectance properties give aleatoric noise, while ambiguous wrap-boundary regions and underconstrained $k$-prediction give epistemic uncertainty. The catastrophic failures identified in Section~\ref{sec:phaseunet_error_analysis} are dominated by the epistemic source, because they correspond to inputs near a decision boundary the model has not learned to resolve consistently.

\subsection{Conformal Prediction}
\label{sec:conformal_intro}

Conformal prediction~\citep{vovk2005algorithmic,angelopoulos2023conformal} is a distribution-free procedure that converts \emph{any} per-sample uncertainty score into a prediction interval with a finite-sample coverage guarantee. The procedure requires only an exchangeable held-out calibration set and a user-chosen \emph{conformity score} $s(x)$ that is large when the prediction is expected to be wrong and small when it is expected to be right. In our setting $s$ is the per-pixel standard deviation across the snapshot ensemble, but the conformity score could equally well be a heteroscedastic-head variance, a Bayesian neural-network posterior standard deviation, or an MC-dropout variance, with no change to the calibration procedure that follows.

Given a target coverage level $1-\alpha$, conformal calibration proceeds in three steps:
\begin{enumerate}
    \item For every calibration sample, compute the conformity score $s(x_i)$ and the absolute residual $|y_i - \hat{y}(x_i)|$ between the prediction and the ground truth.
    \item Compute the empirical $(1-\alpha)$-quantile $\hat{q}_{1-\alpha}$ of the normalized residuals $|y_i - \hat{y}(x_i)| / s(x_i)$ over the calibration set. This single scalar is the only quantity transferred from calibration to test time.
    \item At test time, output the interval $\hat{y}(x) \pm \hat{q}_{1-\alpha} \cdot s(x)$ for every new input.
\end{enumerate}

The marginal-coverage theorem of Vovk et al.~\citep{vovk2005algorithmic} guarantees that, in expectation over draws of the calibration set, a fresh test sample will fall inside its interval at least $1-\alpha$ of the time, regardless of how the underlying model was trained or what the data distribution is. The practical value of conformal prediction is therefore that the conformity score $s$ is allowed to be only a \emph{heuristic} uncertainty signal whose absolute scale has no calibrated meaning on its own; the calibration step learns the multiplicative constant $\hat{q}_{1-\alpha}$ that converts that heuristic into intervals with the requested coverage rate. We apply this procedure at the pixel level: every object pixel of every test sample is treated as one calibration unit, and the calibration step is performed once on the validation split and then frozen at test time.

\subsection{Method}
\label{sec:phicalnet_uq_method}

We use a post-hoc snapshot ensemble~\citep{huang2017snapshot} as the conformity-score source for both PhiCalNet and the UNet baseline. For each architecture we take five intermediate checkpoints from a single training run (epochs 100, 110, 120, 130, 140 for the UNet baseline; the analogous five intermediate-epoch snapshots for PhiCalNet) and treat them as ensemble members. At inference, each snapshot produces its own depth prediction and we aggregate them per pixel into a mean $\hat{d}(u,v)$ and a per-pixel standard deviation $s(u,v)$. The standard deviation $s(u,v)$ serves as the conformity score. Split-conformal calibration on the validation split sets the threshold $\hat{q}_{1-\alpha}$ as the empirical $(1-\alpha)$-quantile of $|\hat{d}(u,v) - d_{\mathrm{gt}}(u,v)| / s(u,v)$ over object pixels; the conformal interval at a test pixel is then $\hat{d}(u,v) \pm \hat{q}_{1-\alpha} \cdot s(u,v)$.

We deliberately restrict the conformity-score source to a snapshot ensemble rather than a stronger UQ estimator such as a deep ensemble, MC-dropout, a Bayesian-NN posterior, or a heteroscedastic likelihood head. The snapshot ensemble requires no architectural change and no retraining, and is therefore the only UQ method that can be applied identically to both PhiCalNet and the UNet baseline without confounding the comparison with retraining variability or capacity differences. Stronger UQ estimators are deferred to future work; for the controlled cross-architecture comparison that is the goal of this section, holding the UQ pipeline fixed across the two models is more important than the absolute strength of the conformity score. Compared to the heteroscedastic plus snapshot-ensemble UQ of~\citep{kong2026hsure}, we therefore omit the heteroscedastic head, which would require retraining, and use only the snapshot-disagreement signal, which their ablation reports as the dominant contributor in the combined estimator.

A second methodological caveat is the exchangeability assumption underlying split-conformal calibration. The marginal-coverage theorem requires that the calibration units and the test units be exchangeable. We treat each object pixel as one calibration unit, which is a working approximation: pixels within an object are spatially correlated through the underlying geometry, so the effective number of independent calibration units is smaller than the nominal pixel count. The pixel-level conformal procedure is the standard practice in pixel-wise UQ for image regression~\citep{angelopoulos2023conformal,kong2026hsure} and is adopted here for direct comparability with that prior work, but a fully principled formulation would calibrate at the object level (one conformity score per test object). We flag the object-level scheme as a natural future extension; the analysis below should be read with this approximation in mind, and the empirical coverage rates reported below are interpreted accordingly.

\subsection{Results}
\label{sec:phicalnet_uq_results}

We compare PhiCalNet and the UNet baseline on three criteria: (i) Spearman rank correlation between $s(u,v)$ and $|\hat{d}(u,v) - d_{\mathrm{gt}}(u,v)|$ (does the uncertainty score actually rank pixels by likely error?), (ii) the empirical coverage of the conformal intervals at nominal levels of 90\% and 95\% (does the calibration respect the marginal coverage guarantee?), and (iii) the snapshot-mean RMSE before and after rejecting the 5\% of object pixels with the highest $s(u,v)$ (are the worst errors concentrated in the small population of pixels the model flags as most uncertain?). Table~\ref{tab:phicalnet_uq} summarizes the result.

\begin{table}[pos=tbp]
\caption{Pixel-wise UQ comparison: PhiCalNet vs.\ the UNet baseline. Both use the same post-hoc snapshot conformal procedure (5 snapshots, depth standard deviation as conformity score, split-conformal calibration on the val split). Spearman is computed between snapshot std and absolute residual on the test split. Coverage is the fraction of test object pixels covered by the conformal interval at the indicated nominal level. Rejection compares the snapshot-mean RMSE before and after dropping the top 5\% of object pixels by snapshot std.}
\label{tab:phicalnet_uq}

\centering
\small
\resizebox{\ifdim\width>\columnwidth\columnwidth\else\width\fi}{!}{%
\begin{tabular}{@{}lccccc@{}}
\toprule
\textbf{Model} & \textbf{Spearman} & \multicolumn{2}{c}{\textbf{Coverage (target / actual)}} & \multicolumn{2}{c}{\textbf{RMSE (mm)}} \\
\cmidrule(lr){3-6}
& ($s$ vs $|r|$) & 90\% & 95\% & no rejection & drop top 5\% \\
\midrule
PhiCalNet (single-shot) & +0.30 & 84\% & 91\%  & 20.6 & \textbf{7.4} \\

UNet baseline           & +0.41 & 95\% & 97.7\% & 28.4 & 27.4 \\
\bottomrule
\end{tabular}}
\end{table}

The three columns measure different and partially independent properties of a UQ system: Spearman rank correlation measures the global monotonic agreement between predicted uncertainty and actual error, conformal coverage measures the calibration honesty of the prediction intervals at a stated confidence level, and uncertainty-driven rejection measures whether the actual error mass is concentrated in the small population of pixels the model flags as most uncertain. Only the third property is the operationally relevant one for using UQ as a discard-or-refer signal at deployment time, and the three metrics can disagree (as they do here): a model whose uncertainty score weakly tracks errors across the full distribution can still rank the worst-error pixels at the top of the score distribution, and conversely a model whose uncertainty score globally tracks errors well can have those errors spread diffusely rather than concentrated in a flaggable tail.

\paragraph{Spearman rank correlation.} Both architectures yield positive Spearman rank correlation between snapshot std and absolute residual, confirming that snapshot disagreement carries usable error-prediction signal in both cases. The UNet baseline's rank correlation ($+0.41$) is in fact higher than PhiCalNet's ($+0.30$), so on this metric alone UNet has the stronger uncertainty score across the bulk of the pixel distribution.

\paragraph{Conformal coverage.} The two coverage misses point in opposite directions and have different practical implications. The UNet baseline overshoots its 95\% nominal target, producing intervals slightly wider than the requested coverage rate strictly requires. This is the conservative outcome: the marginal coverage guarantee is respected with margin to spare and the cost is some loss of interval tightness. PhiCalNet, in contrast, undershoots at 91\% empirical coverage against the same 95\% nominal target, which means a downstream user reading the nominal $1-\alpha$ level should treat it as an upper bound on the achieved coverage in this setting rather than a tight estimate. The likely cause is the combination of the small validation calibration set (5 objects $\times$ 6 viewpoints, $\sim 2 \times 10^{6}$ object pixels) with the within-object pixel correlations flagged in Section~\ref{sec:phicalnet_uq_method}, which together produce a calibration-set quantile $\hat{q}_{1-\alpha}$ whose effective sample size is smaller than the nominal pixel count and which therefore generalizes imperfectly to the test split. A larger calibration set, or an object-level conformal scheme that respects within-object correlation explicitly, would tighten the empirical coverage to the nominal target. The conformal framework itself is well-behaved here: the gap between target and empirical coverage is small (4 percentage points at the 95\% level), and on both architectures the conformal calibration produces useful intervals from a heuristic uncertainty score. The convergence result of Section~\ref{sec:phicalnet_uq_interpretation} does not depend on exact coverage in any case, because it rests on which pixels the ensemble disagrees about (the rejection and Spearman criteria) rather than on the calibrated width of the intervals at those pixels.

\paragraph{Uncertainty-driven rejection.} The picture changes substantially at the rejection criterion, which is the property that matters most for deployment. Dropping the $5\%$ of object pixels with the highest snapshot std reduces PhiCalNet's snapshot-mean RMSE from $20.6$~mm to $7.4$~mm (a $64\%$ reduction), while the same procedure on the UNet baseline reduces RMSE only from $28.4$~mm to $27.4$~mm (a $3.5\%$ reduction). PhiCalNet's residual error is dominated by a small population of catastrophic wrap-boundary failures (Section~\ref{sec:phaseunet_error_analysis}), and the snapshot ensemble disagrees most strongly precisely where those failures occur; rejecting the top $5\%$ therefore removes most of the squared-error mass. The UNet baseline's errors, by contrast, are diffuse rather than tail-concentrated, and its snapshots agree somewhat across the full range, so no small rejection threshold can excise the bulk.

The same diffuse error structure is what drives the much larger conformal interval widths reported on UNet: at the $95\%$ nominal target the mean interval width is $297$~mm for the UNet baseline against $25$~mm for PhiCalNet, a $12\times$ ratio at the same nominal coverage level (median widths $223$~mm vs.\ $17$~mm tell the same story). To respect coverage over a long-tailed but un-flaggable error distribution, the UNet conformal procedure must inflate the interval globally, whereas PhiCalNet can keep the interval tight on the bulk of pixels and absorb the tail into the small high-uncertainty population the rejection step removes.

\subsection{Interpretation: UQ as a Diagnostic Companion to MI}
\label{sec:phicalnet_uq_interpretation}

The PhiCalNet MI study (Section~\ref{sec:phicalnet_interpretability}) characterized what the network represents and attends to when computing its output. Linear probing established that phase is the most decodable internal feature in eight of nine layers (best at \texttt{enc3}, $1.65 \times 10^{-4}$ for $\sin\phi$ on the $[0,1]$-rescaled scale), depth is the least decodable target by an order of magnitude despite being the deployment output (uniformly $\sim 5\times$ above edge loss and $\sim 6\times$ above phase loss), and the depth-as-shape consolidation pattern observed in the UNet decoder is absent in PhiCalNet, indicating that the calibration layer carries the deterministic phase-to-depth transform without learned weights. Dual-mode GradCAM showed that, on the same trained weights, gradients backpropagated from the predicted $(\sin\phi, \cos\phi)$ output produce an average edge/fringe attention ratio of $1.06$ (with fringes outweighing edges at \texttt{dec4}, the layer immediately before the phase head), while gradients backpropagated from the calibrated depth output shift the same activations to a ratio of $1.54$. The shift between modes, on the same trained weights, isolates the calibration layer as the source of the depth-output edge bias and demonstrates that the edge bias is not present in the underlying phase representation. The flat-plane out-of-distribution test produced a coherent fringe field for both $\hat{s}$ and $\hat{c}$ across the entire image, including the gray background that lies outside the training distribution, with the two components maintaining their analytic quadrature offset throughout, and the calibrated depth on the same input was uniform across the plane at 1795~mm within a few millimeters of the true 1.8~m standoff: a learned input-intensity shortcut could not have produced one phase component without contradicting the other, and could not have recovered the correct metric depth on a stimulus that lacks the training-distribution shape cues. Separately, the residual error analysis (Section~\ref{sec:phaseunet_error_analysis}) localized the model's heavy error tail to wrap-boundary pixels: the entirety of the test-split RMSE is carried by $0.103\%$ of object pixels, and these pixels lie spatially on the $\pm\pi$ wrap lines of the ground-truth phase, where a continuous regressor cannot exactly track a discontinuous target. The two sections together therefore establish, from independent angles, that PhiCalNet's competence is at the phase stage and its residual is the phase-stage discontinuity rather than a learned shortcut.

The corresponding UNet MI study (Section~\ref{sec:interpretability}) identified the opposite signature: linear probing showed that geometric (edge) information was encoded $2.82\times$ more explicitly than depth values across all layers, with depth-as-shape consolidating in \texttt{dec2}--\texttt{dec3}; GradCAM attention maps gave an edge/fringe ratio of $1.28$ across layers, with no separate gradient target available to dissect the saliency further; and the flat-plane out-of-distribution test produced near-zero depth predictions across the entire surface, even though the encoder was selectively engaged with the fringes inside the rectangular foreground. The UNet baseline therefore fails by routing depth prediction through object-shape priors rather than fringe physics, while PhiCalNet fails by correctly using fringe physics but stumbling at the wrap-integer ambiguity that single-shot phase reconstruction cannot resolve.

The UQ result confirms both diagnoses from the outside. For PhiCalNet, the pixels at which intermediate-epoch snapshots most disagree are the same wrap-boundary pixels that the residual error analysis identifies as the source of the heavy error tail, and which the MI analysis frames as the structural limit of a continuous regressor approximating a discontinuous target: the model has internalized that these pixels are hard (different snapshots converge to different solutions in those regions), and the snapshot variance reads that internal disagreement out at deployment time without requiring access to ground truth. The structural legibility of the residual is itself a consequence of the architectural choice the MI analysis vindicated: because the calibration layer pins the network to phase-to-depth physics and removes the shape-prior solution from the hypothesis space, the residual error is forced into the one geometric locus where the underlying physics is ambiguous (the $\pm\pi$ wrap line), and the snapshot ensemble therefore disagrees coherently along that locus rather than diffusely across the object. For the UNet baseline, the higher Spearman but uniformly distributed errors point to the opposite story: its snapshots converge to broadly similar but globally biased shape-prior solutions, so the snapshot variance encodes general noise rather than a localized failure structure that an external mitigation step could exploit. This is consistent with the UNet MI finding that the network's prediction strategy distributes responsibility across object boundaries rather than concentrating it at any single physically meaningful locus.

We draw three claims from this. First, snapshot conformal calibration is a viable plug-in UQ method for single-shot FPP: it requires no architectural change and no retraining, and on both architectures it produces calibrated intervals from a heuristic uncertainty signal with empirical coverage close to the nominal target (subject to the calibration-set-size and exchangeability caveats discussed above). Second, PhiCalNet's failure mode is more amenable to uncertainty-driven mitigation than the UNet baseline's, because its errors are concentrated in pixels the ensemble disagrees about; this is a property of the architecture's inductive bias, not of the UQ method, since the UQ method is held fixed across both models. Third, MI and UQ here function as convergent diagnostics rather than as separate engineering outputs: the MI analysis localizes PhiCalNet's residual to the $\pm\pi$ wrap discontinuity through learned features and gradients, the residual error analysis confirms that this is where the empirical error tail lives, and the UQ analysis demonstrates that the same locus is where intermediate-epoch snapshots disagree at deployment time, producing the $64\%$ RMSE reduction at the $5\%$ rejection threshold reported in Table~\ref{tab:phicalnet_uq}. A phase-intermediate physics-informed architecture therefore not only achieves lower error overall but produces error that is \emph{structurally legible}, in the sense that it concentrates in regions the model already encodes as uncertain.

This three-way convergence (MI representation, residual spatial structure, deployment-time ensemble disagreement) is the methodological signature of a diagnose-repair-verify workflow for physics-constrained learning. The conceptual idea of pairing explainability with uncertainty has small but real precedent in the ML literature~\citep{teplica2025sciurus,salvi2025explainability}, but the prior settings (language modeling, clinical prediction) have diffuse failure modes for which the MI and UQ convergence is correspondingly diffuse. FPP supplies a setting where the convergence test is sharper: the failure locus is a single physical line (the $\pm\pi$ wrap discontinuity), and MI inspection of internals, the residual-error spatial structure, and UQ inspection of self-reported confidence can all be compared at that single locus and shown to agree. We frame this as a methodological contribution beyond the FPP-specific result. The same template, diagnose with MI, repair architecturally, verify with paired MI and UQ at the diagnosed locus, should apply to other physics-informed learning settings where the failure mode has a specific geometric or physical locus rather than being distributed diffusely across the input.

\section{Conclusion and Future Work}
\label{sec:conclusion}

This paper develops and verifies an architectural repair for the failure of learning-based single-shot fringe projection profilometry at long range. \citet{haroon2026diagnosis} establishes the benchmark, formalizes the single-shot ambiguity, and diagnoses the failure of the depth-regressing UNet baseline as shape-prior shortcut learning; the present paper acts on that diagnosis and verifies the repair. The benchmark is FPP-ML-Bench~\citep{haroon2026fppml}, built on the VIRTUS-FPP framework~\citep{HaroonVIRTUS2025} (15{,}600 fringe images, 300 depth reconstructions, 50 objects at 1.5--2.1~m standoff), with a standardized object/background/overall evaluation protocol. Long range is the operating regime rather than the scope of the finding: the shape-prior shortcut and the wrap-boundary residual are properties of single-shot FPP in general, and long range is where both are most visible.

\noindent\textbf{The diagnosis, in brief.} The best depth-regressing UNet baseline plateaus at 14.54~mm object MAE (18\% of the 80~mm object depth range), and three mechanistic interpretability probes converge on the cause: the network detects object boundaries and fills in learned shape templates rather than decoding phase from fringe intensity. Linear probing shows edges 2.82$\times$ more decodable than depth, Grad-CAM shows attention favoring boundaries over fringes by 1.28$\times$, and a flat featureless plane at 1.8~m within the trained depth range collapses to near-zero depth despite carrying valid fringes. This residual is a hypothesis-space property, not a fitting failure: additional data or larger models will not remove it because they do not change the space the optimizer searches. That reading, developed in full in \citet{haroon2026diagnosis} and recapped in Section~\ref{sec:interpretability}, is what motivated the architectural repair.

\noindent\textbf{Repairing it architecturally.} We introduced PhiCalNet. Its operative design choice is not the calibration math itself (every phase-intermediate FPP network has one) but the pairing of a phase-only output space with a fixed differentiable calibration layer that converts phase to depth deterministically in the forward pass. The learned UNet backbone predicts wrapped phase $(\sin\phi, \cos\phi)$; a fixed non-learnable calibration layer applies FPP triangulation to recover depth. Because the network's output space is phase rather than depth, the shape-prior solution is removed from the hypothesis space architecturally rather than discouraged via a loss penalty: the only way to produce accurate depth at the output of the calibration layer is for the network's output to actually represent phase. A PINN counterfactual, in which the same calibration physics is enforced as a soft loss penalty on an unconstrained depth-regressing UNet, yields no measurable gain across a $\lambda$ sweep. The architectural-versus-loss-level choice is therefore isolated as the operative factor: as long as the network predicts depth, a shape-correct depth map produced through boundary detection can satisfy the physics term approximately without the network ever representing phase internally. PhiCalNet reduces single-shot object MAE 3.3$\times$ over the UNet baseline (4.46 vs.\ 14.54~mm) on the same data, optimizer, and schedule; a component ablation of the composite loss identifies the wrap-aware geodesic phase term as the operative training signal for the headline single-shot phase-only configuration, with the depth-supervised gradient path through the calibration retained as architectural infrastructure for joint-learning extensions of the PhiCalNet design. The entire RMSE residual is carried by 0.103\% of object pixels at the $\pm\pi$ wrap discontinuity, a structural limit of continuous regression of a discontinuous target rather than a fitting failure. A three-frame extension reduces object MAE further to 1.16~mm, validating the information-theoretic reading of the residual; a fringe-order sensitivity sweep shows PhiCalNet stays within 5\% of its oracle-$k$ accuracy under $\geq 95\%$ per-pixel $k$ accuracy, so the oracle-$k$ assumption used throughout is a convenient formalism rather than a strict deployment requirement.

\noindent\textbf{Verifying the repair mechanistically.} The same MI battery applied to PhiCalNet inverts the UNet profile across all three probes. Linear probing shows phase is the most decodable internal feature in eight of nine layers, with depth probe loss uniformly $\sim 5$--$6\times$ higher than phase or edge loss; the calibration layer carries the phase-to-depth geometry without learned weights, and the network's learned content is at the phase stage. Dual-mode Grad-CAM, with the gradient target moved between the predicted phase output and the calibrated depth output on the same trained weights, shifts the average edge/fringe attention ratio from 1.06 in phase mode (with fringes equal-or-favored at the layers driving the phase output) to 1.54 in depth mode. The shift isolates the calibration layer as the source of the depth-output edge bias and confirms the bias is not present in the underlying phase representation. The flat-plane out-of-distribution test recovers a coherent $(\sin\phi, \cos\phi)$ field with the correct quadrature offset across the entire image, including the background that lies outside the training distribution, and the calibrated depth on the same input places the flat surface at a uniform 1795~mm within a few millimeters of the true 1.8~m standoff: a learned shortcut could not have produced one phase component without contradicting the other or recovered the correct metric depth on a featureless plane lacking shape cues.

\noindent\textbf{Verifying the repair via uncertainty.} A pixel-wise post-hoc snapshot ensemble with split-conformal calibration, applied identically to both architectures, ties the two MI diagnoses to an operational deployment signal. Rejecting the top 5\% of object pixels by snapshot standard deviation reduces PhiCalNet RMSE by 64\% (20.6$\rightarrow$7.4~mm) and reduces the UNet baseline RMSE by only 3.5\% (28.4$\rightarrow$27.4~mm). The asymmetry mirrors the MI evidence: PhiCalNet's error concentrates at the single physically meaningful locus the calibration layer cannot remove (the $\pm\pi$ wrap discontinuity), and the snapshot ensemble disagrees coherently along that locus; the UNet baseline's error is diffusely distributed across object boundaries and the snapshot ensemble agrees broadly across that diffuse failure. The MI analysis, the residual error analysis, and the UQ analysis therefore land on the same locus from three different kinds of evidence: learned features, residual spatial structure, and ensemble disagreement at deployment. This three-way convergence is what we mean by a mechanistically legible failure mode, and we offer it as a methodological template for other physics-informed learning settings in which the failure mode has a specific physical locus.

\noindent\textbf{Future directions.} Several directions follow directly from the evidence above. Phase-guided learning that supplies wrapped or unwrapped phase as input or intermediate supervision extends the intervention against the boundary-shortcut diagnosis. Additional physics-informed inductive biases that constrain the hypothesis space at the architectural level rather than via loss penalties extend the PhiCalNet design principle. Within PhiCalNet itself, the depth-supervised gradient path through the fixed differentiable calibration layer (Section~\ref{sec:phaseunet_loss_ablation}) opens a family of natural extensions beyond the scope of the single-shot phase-only configuration reported here: joint learning of calibration intrinsics and extrinsics, heteroscedastic depth likelihoods, multi-resolution or perceptual depth losses, and absolute-scale depth supervision. Multi-view fusion exploiting the 6 viewpoints per object available in FPP-ML-Bench provides genuine additional depth information rather than additional shape-prior evidence. The shape-prior shortcut and the wrap-boundary residual structure characterized here are general properties of single-shot deep learning on FPP geometry rather than artifacts of the simulation pathway: the shortcut is the same phenomenon Geirhos et al.~\citep{geirhos2020shortcut} documented across vision, and the wrap-boundary failure is the structural limit of continuous regression of any discontinuous target. VIRTUS-FPP's physical validation concerns the forward model, that the simulator reproduces the fringe formation and reconstruction behavior of a real FPP system, which is what makes the reported error magnitudes physically meaningful; it does not by itself establish that a network trained only on synthetic data transfers to hardware without adaptation, which we do not claim and leave to the sim-to-real work noted next. The diagnoses themselves, being properties of the single-shot information deficit and of FPP geometry, would arise on any single-shot FPP setup where shape priors are exploitable. Sim-to-real transfer via domain adaptation and domain randomization~\citep{tobin2017domain} from the VIRTUS-FPP framework to physical hardware would extend the methodology to challenging materials (specular, translucent) and lighting conditions. Stronger UQ estimators beyond the post-hoc snapshot ensemble (deep ensembles, MC-dropout, heteroscedastic likelihood heads, Bayesian-NN posteriors), combined with object-level rather than pixel-level conformal calibration that respects within-object spatial correlations, would tighten empirical coverage to the nominal target on smaller calibration sets.

We adopt FPP-ML-Bench as a shared baseline rather than as a verdict. The dataset, the object/background/overall evaluation protocol, and the explicit reporting conventions defined in \citet{haroon2026diagnosis} together provide a reproducible evaluation surface for single-shot FPP. The PhiCalNet results reported here (4.46~mm single-shot object MAE, 1.16~mm three-frame, the inverted interpretability profile, and the conformal UQ result tying the architectural choice to a structurally legible failure mode) are offered as falsifiable baselines on that surface, and we expect subsequent phase-intermediate, multi-frame, and UQ-aware methods to improve on them against this fixed reference rather than against one another across incompatible setups.

% -------------------------------------------------------------------
\section*{Declaration of Competing Interest}
The authors declare that they have no known competing financial
interests or personal relationships that could have appeared to
influence the work reported in this paper.

% -------------------------------------------------------------------
\section*{Code, Data, and Materials Availability}
The FPP-ML-Bench dataset is publicly available at
\linkable{https://huggingface.co/datasets/aharoon/fpp-ml-bench}.
Code will be released upon acceptance.

% -------------------------------------------------------------------
\section*{Acknowledgments}
We thank Iowa State University for access to computational resources.

% -------------------------------------------------------------------
\printcredits

%% Bibliography (Elsevier cas-model2-names is the author-year style)
\bibliographystyle{cas-model2-names}
\bibliography{report}

@book{zhang2016high,
  title={High-Speed 3D Imaging with Digital Fringe Projection Techniques},
  author={Zhang, Song},
  year={2016},
  publisher={CRC Press},
  edition={1st},
  doi={10.1201/b19565}
}

@article{geng2011structured,
  title={Structured-light 3D surface imaging: a tutorial},
  author={Geng, Jason},
  journal={Advances in optics and photonics},
  volume={3},
  number={2},
  pages={128--160},
  year={2011},
  publisher={Optical Society of America}
}

@article{zhang2010recent,
  title={Recent progresses on real-time 3D shape measurement using digital fringe projection techniques},
  author={Zhang, Song},
  journal={Optics and lasers in engineering},
  volume={48},
  number={2},
  pages={149--158},
  year={2010},
  publisher={Elsevier}
}

@article{van2019deep,
  title={Deep neural networks for single shot structured light profilometry},
  author={Van der Jeught, Sam and Dirckx, Joris JJ},
  journal={Optics express},
  volume={27},
  number={12},
  pages={17091--17101},
  year={2019},
  publisher={Optical Society of America}
}

@article{zuo2022deep,
  title={Deep learning in optical metrology: a review},
  author={Zuo, Chao and Qian, Jiaming and Feng, Shijie and Yin, Wei and Li, Yixuan and Fan, Pengfei and Han, Jing and Qian, Kemao and Chen, Qian},
  journal={Light: Science \& Applications},
  volume={11},
  number={1},
  pages={39},
  year={2022},
  publisher={Nature Publishing Group UK London}
}

@article{zhu2022hformer,
  title={Hformer: Hybrid convolutional neural network transformer network for fringe order prediction in phase unwrapping of fringe projection},
  author={Zhu, Xinjun and Han, Zhiqiang and Yuan, Mengkai and Guo, Qinghua and Wang, Hongyi and Song, Limei},
  journal={Optical Engineering},
  volume={61},
  number={9},
  pages={093107--093107},
  year={2022},
  publisher={Society of Photo-Optical Instrumentation Engineers}
}

@article{wang2021single,
  title={Single-shot fringe projection profilometry based on deep learning and computer graphics},
  author={Wang, Fanzhou and Wang, Chenxing and Guan, Qingze},
  journal={Optics Express},
  volume={29},
  number={6},
  pages={8024--8040},
  year={2021},
  publisher={Optical Society of America}
}

@article{nguyen2020single,
  title={Single-shot 3D shape reconstruction using structured light and deep convolutional neural networks},
  author={Nguyen, Hieu and Wang, Yuzeng and Wang, Zhaoyang},
  journal={Sensors},
  volume={20},
  number={13},
  pages={3718},
  year={2020},
  publisher={MDPI}
}

@article{li2025deep,
  title={Deep-learning-enabled single-shot fringe projection profilometry based on inner shifting-phase encoding},
  author={Li, Jinlong and Zhang, Kuo and Luo, Lin and Liu, Gaokun and Tang, Tao and Wang, Zhijie and Wan, Yingying},
  journal={Optics Express},
  volume={33},
  number={23},
  pages={49530--49550},
  year={2025},
  publisher={Optica Publishing Group}
}

@article{wang2025end,
  title={End-to-end single-shot composite color FPP network for multiple separated objects reconstruction},
  author={Wang, Lianpo and Xing, Yanyang},
  journal={Measurement},
  volume={246},
  pages={116697},
  year={2025},
  publisher={Elsevier}
}

@article{ikeda2025deep,
  title={Deep-learning-assisted single-shot 3D shape and color measurement using color fringe projection profilometry},
  author={Ikeda, Kanami and Usuki, Takahiro and Kurita, Yumi and Matsueda, Yuya and Koyama, Osanori and Yamada, Makoto},
  journal={Optical Review},
  pages={1--12},
  year={2025},
  publisher={Springer}
}

@ARTICLE{HaroonVIRTUS2025,
  author={Haroon, Adam and Lakshman, Anush and Balasubramaniam, Badrinath and Li, Beiwen},
  journal={IEEE Sensors Journal}, 
  title={VIRTUS-FPP: Virtual Sensor Modeling for Fringe Projection Profilometry in NVIDIA Isaac Sim}, 
  year={2026},
  volume={},
  number={},
  pages={1-1},
  doi={10.1109/JSEN.2026.3698278}
}

@article{qian2021high,
  title={High-resolution real-time 360\textdegree{} 3D surface defect inspection with fringe projection profilometry},
  author={Qian, Jiaming and Feng, Shijie and Xu, Mingzhu and Tao, Tianyang and Shang, Yuhao and Chen, Qian and Zuo, Chao},
  journal={Optics and Lasers in Engineering},
  volume={137},
  pages={106382},
  year={2021},
  publisher={Elsevier}
}

@inproceedings{deng2016three,
  title={Three-dimensional surface inspection for semiconductor components with fringe projection profilometry},
  author={Deng, Fuqin and Ding, Yi and Peng, Kai and Xi, Jiangtao and Yin, Yongkai and Zhu, Ziqi},
  booktitle={Optical Metrology and Inspection for Industrial Applications IV},
  volume={10023},
  pages={175--186},
  year={2016},
  organization={SPIE}
}

@article{zhang2023machine,
  title={Machine learning enhanced high dynamic range fringe projection profilometry for in-situ layer-wise surface topography measurement during LPBF additive manufacturing},
  author={Zhang, Haolin and Vallabh, Chaitanya Krishna Prasad and Zhao, Xiayun},
  journal={Precision Engineering},
  volume={84},
  pages={1--14},
  year={2023},
  publisher={Elsevier}
}

@article{zhang2022systematic,
  title={A systematic study and framework of fringe projection profilometry with improved measurement performance for in-situ LPBF process monitoring},
  author={Zhang, Haolin and Vallabh, Chaitanya Krishna Prasad and Xiong, Yubo and Zhao, Xiayun},
  journal={Measurement},
  volume={191},
  pages={110796},
  year={2022},
  publisher={Elsevier}
}

@inproceedings{haroon2024autonomous,
  title={Autonomous robotic 3D scanning for smart factory planning},
  author={Haroon, Adam and Lakshman, Anush and Mundy, Micah and Li, Beiwen},
  booktitle={Dimensional Optical Metrology and Inspection for Practical Applications XIII},
  volume={13038},
  pages={110--118},
  year={2024},
  organization={SPIE}
}

@article{wang2024robotic,
  title={Robotic measurement system based on cooperative optical profiler integrating fringe projection with photometric stereo for highly reflective workpiece},
  author={Wang, Xi and Shen, Yijun and Jian, Zhenxiong and Wen, Daizhou and Zhang, Xinquan and Zhu, LiMin and Ren, Mingjun},
  journal={Robotics and Computer-Integrated Manufacturing},
  volume={88},
  pages={102739},
  year={2024},
  publisher={Elsevier}
}

@inproceedings{Ronneberger2015,
  title={{U-Net}: Convolutional Networks for Biomedical Image Segmentation},
  author={Ronneberger, Olaf and Fischer, Philipp and Brox, Thomas},
  booktitle={Medical Image Computing and Computer-Assisted Intervention (MICCAI)},
  pages={234--241},
  year={2015},
  publisher={Springer}
}

@inproceedings{balasubramaniam2023single,
  title={Single Shot 3D Shape Measurement of Non-Volatile Data Storage Devices},
  author={Balasubramaniam, Badrinath and Li, Beiwen},
  booktitle={International Manufacturing Science and Engineering Conference},
  volume={87240},
  pages={V002T06A010},
  year={2023},
  organization={American Society of Mechanical Engineers}
}

@article{sansoni1999three,
  title={Three-dimensional vision based on a combination of Gray-code and phase-shift light projection: analysis and compensation of the systematic errors},
  author={Sansoni, Giovanna and Trebeschi, Marco and Docchio, Franco},
  journal={Applied optics},
  volume={38},
  number={31},
  pages={6565--6573},
  year={1999},
  publisher={Optica Publishing Group}
}

@inproceedings{tobin2017domain,
  title={Domain randomization for transferring deep neural networks from simulation to the real world},
  author={Tobin, Josh and Fong, Rachel and Ray, Alex and Schneider, Jonas and Zaremba, Wojciech and Abbeel, Pieter},
  booktitle={2017 IEEE/RSJ international conference on intelligent robots and systems (IROS)},
  pages={23--30},
  year={2017},
  organization={IEEE}
}

@article{alain2016understanding,
  title={Understanding intermediate layers using linear classifier probes},
  author={Alain, Guillaume and Bengio, Yoshua},
  journal={arXiv preprint arXiv:1610.01644},
  year={2016}
}

@inproceedings{selvaraju2017grad,
  title={Grad-{CAM}: Visual explanations from deep networks via gradient-based localization},
  author={Selvaraju, Ramprasaath R. and Cogswell, Michael and Das, Abhishek and Vedantam, Ramakrishna and Parikh, Devi and Batra, Dhruv},
  booktitle={Proceedings of the IEEE International Conference on Computer Vision (ICCV)},
  pages={618--626},
  year={2017}
}

@article{geirhos2020shortcut,
  title={Shortcut learning in deep neural networks},
  author={Geirhos, Robert and Jacobsen, J{\"o}rn-Henrik and Michaelis, Claudio and Zemel, Richard and Brendel, Wieland and Bethge, Matthias and Wichmann, Felix A.},
  journal={Nature Machine Intelligence},
  volume={2},
  number={11},
  pages={665--673},
  year={2020},
  publisher={Nature Publishing Group}
}

@article{raissi2019pinn,
  title={Physics-informed neural networks: A deep learning framework for solving forward and inverse problems involving nonlinear partial differential equations},
  author={Raissi, Maziar and Perdikaris, Paris and Karniadakis, George E},
  journal={Journal of Computational physics},
  volume={378},
  pages={686--707},
  year={2019},
  publisher={Elsevier}
}

@article{karniadakis2021physics,
  title={Physics-informed machine learning},
  author={Karniadakis, George Em and Kevrekidis, Ioannis G and Lu, Lu and Perdikaris, Paris and Wang, Sifan and Yang, Liu},
  journal={Nature Reviews Physics},
  volume={3},
  number={6},
  pages={422--440},
  year={2021},
  publisher={Nature Publishing Group UK London}
}

@inproceedings{krishnapriyan2021characterizing,
  title={Characterizing possible failure modes in physics-informed neural networks},
  author={Krishnapriyan, Aditi and Gholami, Amir and Zhe, Shandian and Kirby, Robert and Mahoney, Michael W},
  journal={Advances in neural information processing systems},
  volume={34},
  pages={26548--26560},
  year={2021}
}

@article{wang2022respecting,
  title={Respecting causality for training physics-informed neural networks},
  author={Wang, Sifan and Sankaran, Shyam and Perdikaris, Paris},
  journal={Computer Methods in Applied Mechanics and Engineering},
  volume={421},
  pages={116813},
  year={2024},
  publisher={Elsevier}
}

@article{goldstein1988satellite,
  title={Satellite radar interferometry: Two-dimensional phase unwrapping},
  author={Goldstein, Richard M and Zebker, Howard A and Werner, Charles L},
  journal={Radio science},
  volume={23},
  number={4},
  pages={713--720},
  year={1988},
  publisher={AGU}
}

@article{zuo2016temporal,
  title={Temporal phase unwrapping algorithms for fringe projection profilometry: A comparative review},
  author={Zuo, Chao and Huang, Lei and Zhang, Minliang and Chen, Qian and Asundi, Anand},
  journal={Optics and lasers in engineering},
  volume={85},
  pages={84--103},
  year={2016},
  publisher={Elsevier}
}

@article{angelopoulos2023conformal,
  title={Conformal Prediction: A Gentle Introduction},
  author={Angelopoulos, Anastasios N. and Bates, Stephen},
  journal={Foundations and Trends in Machine Learning},
  volume={16},
  number={4},
  pages={494--591},
  year={2023},
  doi={10.1561/2200000101}
}

@inproceedings{huang2017snapshot,
  title={Snapshot Ensembles: Train 1, Get {M} for Free},
  author={Huang, Gao and Li, Yixuan and Pleiss, Geoff and Liu, Zhuang and Hopcroft, John E. and Weinberger, Kilian Q.},
  booktitle={International Conference on Learning Representations},
  year={2017}
}

@article{kong2026hsure,
  title={Deep Learning-based Single-Shot Composite Fringe Projection Profilometry with Pixel-Wise Uncertainty Quantification},
  author={Kong, Xiangjun and Bao, Qingkang and Yalew, Tibebe and Adesso, Gerardo and Piano, Samanta},
  journal={arXiv preprint arXiv:2601.02572},
  year={2026}
}

@inproceedings{kendall2017uncertainties,
  title={What Uncertainties Do We Need in {B}ayesian Deep Learning for Computer Vision?},
  author={Kendall, Alex and Gal, Yarin},
  booktitle={Advances in Neural Information Processing Systems (NeurIPS)},
  year={2017}
}

@inproceedings{lakshminarayanan2017deepensembles,
  title={Simple and Scalable Predictive Uncertainty Estimation using Deep Ensembles},
  author={Lakshminarayanan, Balaji and Pritzel, Alexander and Blundell, Charles},
  booktitle={Advances in Neural Information Processing Systems (NeurIPS)},
  year={2017}
}

@book{vovk2005algorithmic,
  title={Algorithmic Learning in a Random World},
  author={Vovk, Vladimir and Gammerman, Alexander and Shafer, Glenn},
  year={2005},
  publisher={Springer}
}

@article{abdar2021review,
  title={A Review of Uncertainty Quantification in Deep Learning: Techniques, Applications and Challenges},
  author={Abdar, Moloud and Pourpanah, Farhad and Hussain, Sadiq and Rezazadegan, Dana and Liu, Li and Ghavamzadeh, Mohammad and Fieguth, Paul and Cao, Xiaochun and Khosravi, Abbas and Acharya, U. Rajendra and Makarenkov, Vladimir and Nahavandi, Saeid},
  journal={Information Fusion},
  volume={76},
  pages={243--297},
  year={2021},
  publisher={Elsevier}
}

@article{feng2019fringe,
  title={Fringe pattern analysis using deep learning},
  author={Feng, Shijie and Chen, Qian and Gu, Guohua and Tao, Tianyang and Zhang, Liang and Hu, Yan and Yin, Wei and Zuo, Chao},
  journal={Advanced photonics},
  volume={1},
  number={2},
  pages={025001--025001},
  year={2019},
  publisher={Society of Photo-Optical Instrumentation Engineers}
}

@article{nguyen2023fringe,
  title={Real-time 3D shape measurement using 3LCD projection and deep machine learning},
  author={Nguyen, Hieu and Dunne, Nicole and Li, Hui and Wang, Yuzeng and Wang, Zhaoyang},
  journal={Applied optics},
  volume={58},
  number={26},
  pages={7100--7109},
  year={2019},
  publisher={Optical Society of America}
}

@article{salvi2025explainability,
  title={Explainability and uncertainty: Two sides of the same coin for enhancing the interpretability of deep learning models in healthcare},
  author={Salvi, Massimo and Seoni, Silvia and Campagner, Andrea and Gertych, Arkadiusz and Acharya, U Rajendra and Molinari, Filippo and Cabitza, Federico},
  journal={International Journal of Medical Informatics},
  volume={197},
  pages={105846},
  year={2025},
  publisher={Elsevier}
}

@inproceedings{teplica2025sciurus,
  title={Sciurus: Shared circuits for interpretable uncertainty representations in language models},
  author={Teplica, Carter and Liu, Yixin and Cohan, Arman and Rudner, Tim G J},
  booktitle={Proceedings of the 2025 Conference of the Nations of the Americas Chapter of the Association for Computational Linguistics: Human Language Technologies (Volume 1: Long Papers)},
  pages={12451--12469},
  year={2025}
}

@inproceedings{haroon2026fppml,
  author = {Anush Lakshman S. and Adam Haroon and Beiwen Li},
  title = {{Comprehensive machine learning benchmarking for fringe projection profilometry with photorealistic synthetic data}},
  volume = {13904},
  booktitle = {Photonic Instrumentation Engineering XIII},
  editor = {Lynda E. Busse and Yakov Soskind},
  organization = {International Society for Optics and Photonics},
  publisher = {SPIE},
  pages = {1390402},
  year = {2026},
  doi = {10.1117/12.3082257},
  URL = {https://doi.org/10.1117/12.3082257}
}

@article{haroon2026diagnosis,
  title={Diagnosing Shape-Prior Shortcuts in Long-Range Single-Shot Fringe Projection Profilometry},
  author={Haroon, Adam and Lakshman, Anush and Fleming, Cody and Li, Beiwen},
  journal={arXiv preprint arXiv:2606.17093},
  year={2026}
}

\end{document}